\pdfoutput=1

\documentclass[11pt]{article}

\usepackage[final]{acl}

\usepackage{times}
\usepackage{latexsym}

\usepackage[T1]{fontenc}

\usepackage[utf8]{inputenc}

\usepackage{microtype}

\usepackage{inconsolata}

\usepackage{graphicx}
\usepackage{multirow}

\usepackage{subcaption}
\usepackage{tcolorbox}
\usepackage[colorinlistoftodos,prependcaption,textsize=small]{todonotes}
\usepackage{listings}

\usepackage[table]{xcolor}

\definecolor{lightblue}{RGB}{169,204,227}
\definecolor{midblue}{RGB}{84, 153, 199}
\definecolor{darkblue}{RGB}{36,113,163}

\definecolor{lightred}{RGB}{255, 200, 200}
\definecolor{midred}{RGB}{255, 100, 100}
\definecolor{darkred}{RGB}{200, 0, 0}

\newcommand{\highlightpos}[1]{\cellcolor{lightblue}\textbf{#1}}
\newcommand{\highlightposs}[1]{\cellcolor{midblue}\textbf{#1}}
\newcommand{\highlightposss}[1]{\cellcolor{darkblue}\textbf{#1}}

\newcommand{\highlightneg}[1]{\cellcolor{lightred}\textbf{#1}}
\newcommand{\highlightnegg}[1]{\cellcolor{midred}\textbf{#1}}
\newcommand{\highlightneggg}[1]{\cellcolor{darkred}\textbf{#1}}

\newcommand{\webis}{\textit{webis\_reddit}\ }
\newcommand{\twhm}{\textit{100M\_tweets}\ }
\newcommand{\senatortweets}{\textit{senator\_tweets}\ }
\newcommand{\wikipedia}{\textit{wikipedia}\ }
\newcommand{\redditsubmissions}{\textit{reddit\_submissions}\ }

\title{Recursive Training Loops in LLMs: How training data properties modulate distribution shift in generated data?}

\author{
 \textbf{Grgur Kovač\textsuperscript{* 1}},
 \textbf{Jérémy Perez\textsuperscript{* 1}},
 \textbf{Rémy Portelas\textsuperscript{2}},
\\
 \textbf{Peter Ford Dominey\textsuperscript{3, 4}},
 \textbf{Pierre-Yves Oudeyer\textsuperscript{1}}
\\
\scriptsize{
 \textsuperscript{1}Flowers TEAM, INRIA, FR
 \textsuperscript{2}Ubisoft La Forge, FR
 \textsuperscript{3}INSERM UMR 1093-CAPS, FR
 \textsuperscript{4}Robot Cognition Laboratory, Institute Marey, FR
}
\\
\scriptsize{
 \textsuperscript{*}equal contribution
}
\\
\scriptsize{
   Correspondence: \href{mailto:grgur.kovac@inria.fr}{grgur.kovac@inria.fr}
}
}

\begin{document}
\maketitle
\begin{abstract}
Large language models (LLMs) are increasingly used in the creation of online content, creating feedback loops as subsequent generations of models will be trained on this synthetic data. Such loops were shown to lead to \textit{distribution shifts} - models misrepresenting the true underlying distributions of human data (also called \textit{model collapse}). However, how human data properties affect such shifts remains poorly understood. In this paper, we provide the first empirical examination of the effect of such properties on the outcome of recursive training. We first confirm that using different human datasets leads to distribution shifts of different magnitudes. Through exhaustive manipulation of dataset properties combined with regression analyses, we then identify a set of properties associated with distribution shift magnitudes. Lexical diversity is found to amplify these shifts, while semantic diversity and data quality mitigate them. Furthermore, we find that these influences are highly modular: data scrapped from a given internet domain has little influence on the content generated for another domain. Finally, experiments on political bias reveal that human data properties affect whether the initial bias will be amplified or reduced. Overall, our results portray a novel view, where different parts of internet may undergo different types of distribution shift.
\end{abstract}

\section{Introduction}
Large Language Models (LLMs) are increasingly contributing to the creation of internet content, being used for journalism \citep{llm_news}, coding \cite{llm_code_survey} and generating content on social media \cite{the_rise_of_social_bots}.
The increasing amount of synthetic, LLM-generated data on the internet introduces a precarious feedback loop: LLMs trained on datasets containing synthetic data will themselves generate data that will be used to train future models.
\citet{model_collapse} demonstrated that this process, known as \textbf{recursive training}, can have detrimental effects, causing models to progressively lose information about the true underlying distributions they are intended to approximate.
This results in a gradual change in generated distributions, often accompanied by a reduction in variance.
While this has sometimes been described as \textit{model collapse}, we refer to this mismatch between the true and the generated distribution as \textit{distribution shift}, and limit the term collapse to refer to \textit{detrimental} distribution shifts, such as losses in quality or diversity. 
Such detrimental effects \cite{curious_decline_div} as well as bias amplification \cite{bias_amplification} have been reported in previous work.
Their potential societal consequences make it imperative to better understand the dynamics of recursive training. 

\begin{figure*}
\centering
\includegraphics[width=0.9\linewidth]{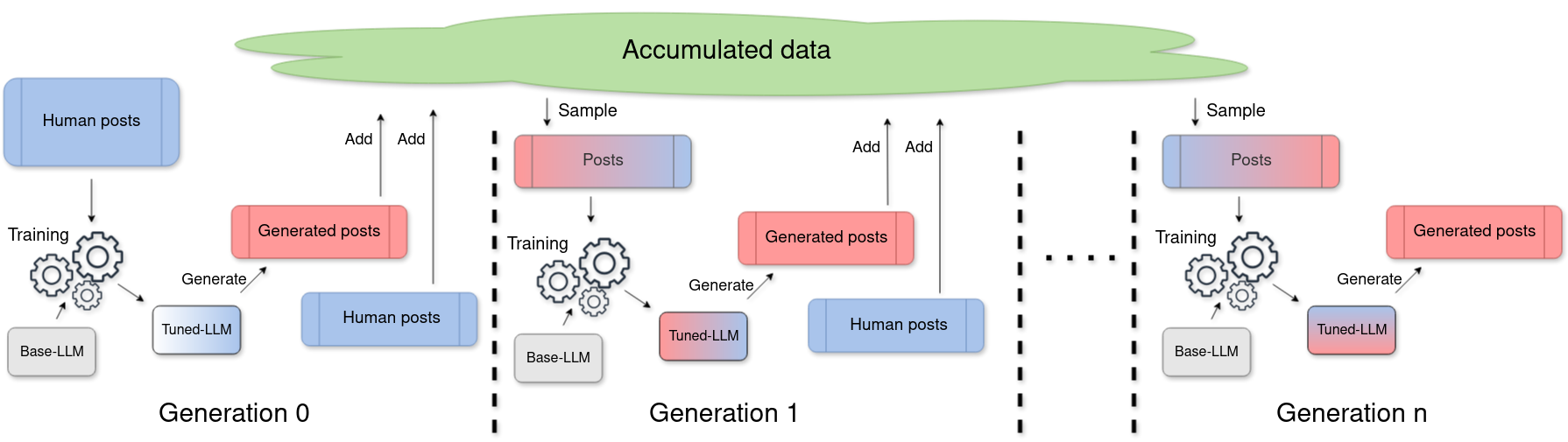}
\caption{\textbf{Iterative chain} In each generation, a fresh base model is fine-tuned on texts sampled from the Accumulated data pool (except generation 0, where it's trained only on human posts). The model generates posts, which are added to the pool alongside some newly sampled human posts.}
\label{fig:iterative_chain}
\end{figure*}

Internet data varies along a wide range of properties.
For instance, certain domains may be associated with a higher ratio of synthetic-to-human data (e.g. GitHub), others with lower quality data (e.g. Reddit), and some with lower diversity data (e.g. specialized forums).
If those properties affects the outcome of recursive fine-tuning, we may expect different types of shifts on different parts of the internet.

Given the early stage of research in this field, existing studies often ignore this diversity, relying on simplifying assumptions and focusing on abstracted settings.
To the best of our knowledge,
how different data properties influence distribution shifts remains largely unexplored, aside from studies examining the ratio of human to synthetic data \citep{Bertrand2023, Bohacek2023, Kazdan2024, Martnez2023, Zhang2024}.
Filing this gap is therefore crucial to draw a more nuanced and detailed picture of the consequences of recursive fine-tuning.

In this paper, we adopt the iterative chain paradigm used in previous studies \citep{model_collapse,accumulation}.
Our experimental setup is shown in Figure~\ref{fig:iterative_chain}.
The process begins with fine-tuning a base LLM on a selection of human data (e.g. Reddit posts).
This model generates data, which are added to the pool of Accumulated data together with newly sampled human data.
In each subsequent generation, a new base model is fine-tuned on data sampled from the Accumulated data pool.
This model, in turn, generates new data, which are again added
to the pool along with fresh human data.
This pipeline allows us to study how generated data evolve across successive generations, with a particular focus on the distribution shift from the first to the final generation.

In our experiments, we study how various properties of human data (e.g. quality, diversity, bias) influence the dynamics of distribution shifts in recursive training chains.
We use five datasets spanning three domains (Twitter, Reddit and Wikipedia).
First, we confirm that the choice of the dataset greatly influences the consequences of iterative fine-tuning (Section \ref{sec:exp_datasets}): while some datasets exhibit sharp decreases in diversity and quality, others are more robust to such shifts. 
Our second set of experiments (Sections \ref{sec:exp_reg} and \ref{sec:exp_reg_merged}) aims to uncover specifically which properties of training data mitigate or amplify distribution shifts.
We run iterative chain experiments on 800 clusters created from four different datasets,
and conduct a series of regression analysis mapping various data properties to the degradation in the quality and diversity of generated texts.
We find that lexical diversity is associated with greater degradation, while semantic diversity has the opposite effect. 
Furthermore, we observe that these influences are highly modular, with generated content being mostly influenced by human data properties from the same domain. This suggests that different internet domains might undergo distinct and relatively independent distribution shifts regardless of models being trained on a mixture of domains. 
Finally, our last set of experiments focuses on the evolution of political bias.
The results indicate that the type of shift observed (bias reduction, amplification or inversion) depends on the political lean of the the human data.
This empirically confirms that properties of human data play an important in shaping the outcome of recursive training.

The code for reproducing the simulations, analyses and figures is available on our GitHub\footnote{https://anonymous.4open.science/r/ce\_llms-9068}.

The main contributions of this work are:
\begin{itemize}
    \item We propose and experimentally confirm the hypothesis that different training datasets lead to different distribution shift dynamics, motivating an investigation on the underlying causes.
    \item Through an extensive set of experiments (four datasets over three domains), we outline several data properties as influencing distribution shift dynamics.
    \item We reveal that these influences are highly modular, with generated content being mostly influenced by human data properties from the same domain.
    \item We find that distribution shifts also occur in terms of political lean, and that the type of shift (bias amplification, reduction or inversion) depends on the political lean of the human data.
\end{itemize}

\section{Related Work}



\paragraph{Recursive fine-tuning and model collapse}
A rapidly growing body of literature has focused on the consequences of recursively training generative models on synthetic data \cite{schaeffer2025position}.
The phrase “model collapse”, coined in \citet{model_collapse}, refers to the progressive degradation of models induced by this feedback loop.
This phenomenon has been studied both theoretically \citep{Dohmatob2024_demist, dohmatob2024tale, Bertrand2023, Alemohammad2023} and empirically, on both generative image models \citep{Martnez2023, Martnez2023b, Bohacek2023, Hataya2022, Alemohammad2023} and language models \cite{Zhang2024, Guo2023, Kazdan2024,Briesch2023,Gerstgrasser2024}.
Theoretical results have provided valuable insights, for instance showing how it is characterized by the disappearance of distribution tails \cite{Dohmatob2024_demist, dohmatob2024tale,model_collapse}.
Empirical studies have allowed to establish several properties of this phenomenon, such as the role of synthetic-to-real-data ratio \citep{Briesch2023, Hataya2022} or strategies for mitigating collapse \citep{Gerstgrasser2024, Kazdan2024, Zhang2024}.
Recently, \citet{Wang2024} showed that recursive LLM fine-tuning can lead to bias amplification. 
These works do not systematically evaluate how the properties of the human dataset used in their experiments affect their conclusions. 
Here, we extend this literature by investigating how those properties impact the outcome of recursive training.

\paragraph{Cultural dynamics in artificial agents}
The motivation for this research area stems from the observation that human-made technologies have transitioned from passive mediators of cultural evolution (e.g., the printing press) to active generators of cultural content.
This shift has been described as the emergence of machine culture - culture mediated or generated by machines \cite{machine_culture}.
Understanding the dynamics that shape the evolution of machine-generated content over time is therefore crucial.
This has led researchers to study cultural dynamics in populations of reinforcement learning agents \citep{cook_artificial_2024, schmitt_kickstarting_2018, open_ended_learning_team_open-ended_2021, prystawski_cultural_2023, nisioti_social_2022} and of LLMs \cite{perez2024llms, perez2024cultural, nisioti2024collective, vallinder2024cultural, burton2024large}.
Our work extends this literature by examining the factors that modulate the evolution of LLM-generated content.

\section{Methods}

\subsection{The iterative chain paradigm}
\label{sec:method_chain}

We use the iterative chain paradigm inspired by \citet{model_collapse,accumulation}.
Our experimental design is shown in Figure~\ref{fig:iterative_chain}.
First, a base LLM is fine-tuned on 8000 samples from a human dataset (e.g. Wikipedia articles or Reddit posts).
This model generates $4000*r$ posts, where $r$ is the synthetic-data ratio.
Those posts are added to the pool of accumulated data together with $4000*(1-r)$ newly sampled human posts.
In all subsequent generations, a new base model is fine-tuned on $4000$ posts sampled from the accumulated data pool, and $4000*r$ posts generated by this model are added to the accumulated data pool together with $4000*(1-r)$ newly sampled human data.
In each generation, a new base model is sampled from four possible options: \texttt{LLaMa-3.2-1B} \citep{llama_3} (LLAMA 3.1 Community), \texttt{Qwen2.5-1.5B} \citep{qwen2.5} (Apache), \texttt{SmolLM-1.7B} \citep{smollm} (Apache), \texttt{Falcon3-1B-Base} \citep{falcon3}), and fine-tuned using LoRA \citep{lora} (see appendix \ref{app:ft} for details).
Five seeds were used in all experiments.
This pipeline enables us to study the evolution of generated data over generations, and most notably, the difference between data generated in the first and last generations.

\subsection{Datasets}

We conducted our experiments on five datasets: two consisting of Twitter posts, two of Reddit posts, and one of Wikipedia paragraphs.
These platforms were chosen because they are likely to be increasingly populated with AI-generated content, often indistinguishable from human-written text.
Additionally, they are frequently scraped to construct training dataset for language models.
Finally, they cover a diverse range of topics and language styles - an essential requirement for investigating the effects of data properties on recursive training dynamics.
Refer to Appendix \ref{app:datasets} for details.

\subsection{Metrics}
\label{sec:metrics}
In this work, we study distribution shift dynamics in terms of quality, semantic diversity and political lean.
Irrespective of how much content a model outputs (which varies with the synthetic-data ratio), we always evaluate those metrics on a sample of 250 generated texts.

\textbf{The Semantic Diversity} of a set of texts is measured as the pairwise cosine diversity in the \texttt{stella\_en\_1.5B\_v5} model \citep{stella}, as in previous studies \citep{curious_decline_div}.
\textbf{Quality} and \textbf{Political lean} are estimated by using \texttt{LLaMa-3.3-70B-Instruct} \citep{llama_3} in the LLM-as-a-Judge setup.
The prompt is inspired by \citet{is_chatgpt_a_good_nlg_evaluator} and \citet{llm_for_reference_free_quality_eval}, and adapted to our task of evaluating the quality of short texts (see appendix \ref{app:llm_judge_validation} or details and validation of that adaptation).
In sections \ref{sec:exp_reg} and \ref{sec:exp_reg_merged}, we additionally rely on the following metrics.
\textbf{Lexical Diversity} is estimated as SelfBLEU \citep{selfbleu}, computed as the average BLEU score \cite{bleu} of each text using all other texts as references, following prior work \cite{curious_decline_div}.
\textbf{Gaussianity} is measured by fitting a 2D UMAP projection on embeddings from the \texttt{stella\_en\_1.5B\_v5} model, and computing the AIC \citep{aic} of a 2D Gaussian distribution fit to this space. 
\textbf{Positivity} is assessed using the \texttt{SentimentIntensityAnalyzer} tool from NLTK \citep{hardeniya2016natural}, which assigns a sentiment score to each text ranging from $-1.0$ (highly negative) to $1.0$ (highly positive).

\begin{figure}
\includegraphics[width=0.49\linewidth]{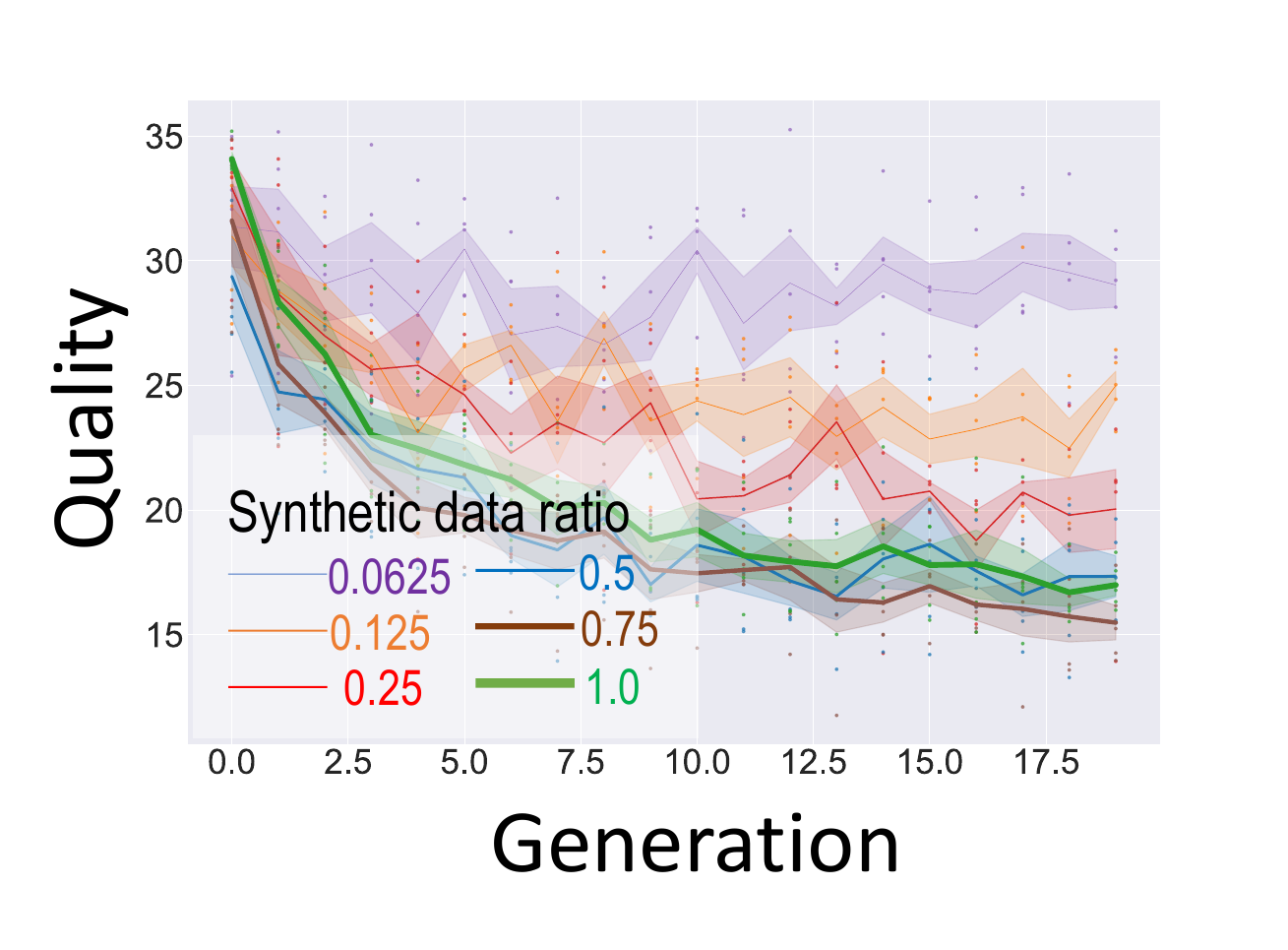}
\includegraphics[width=0.49\linewidth]{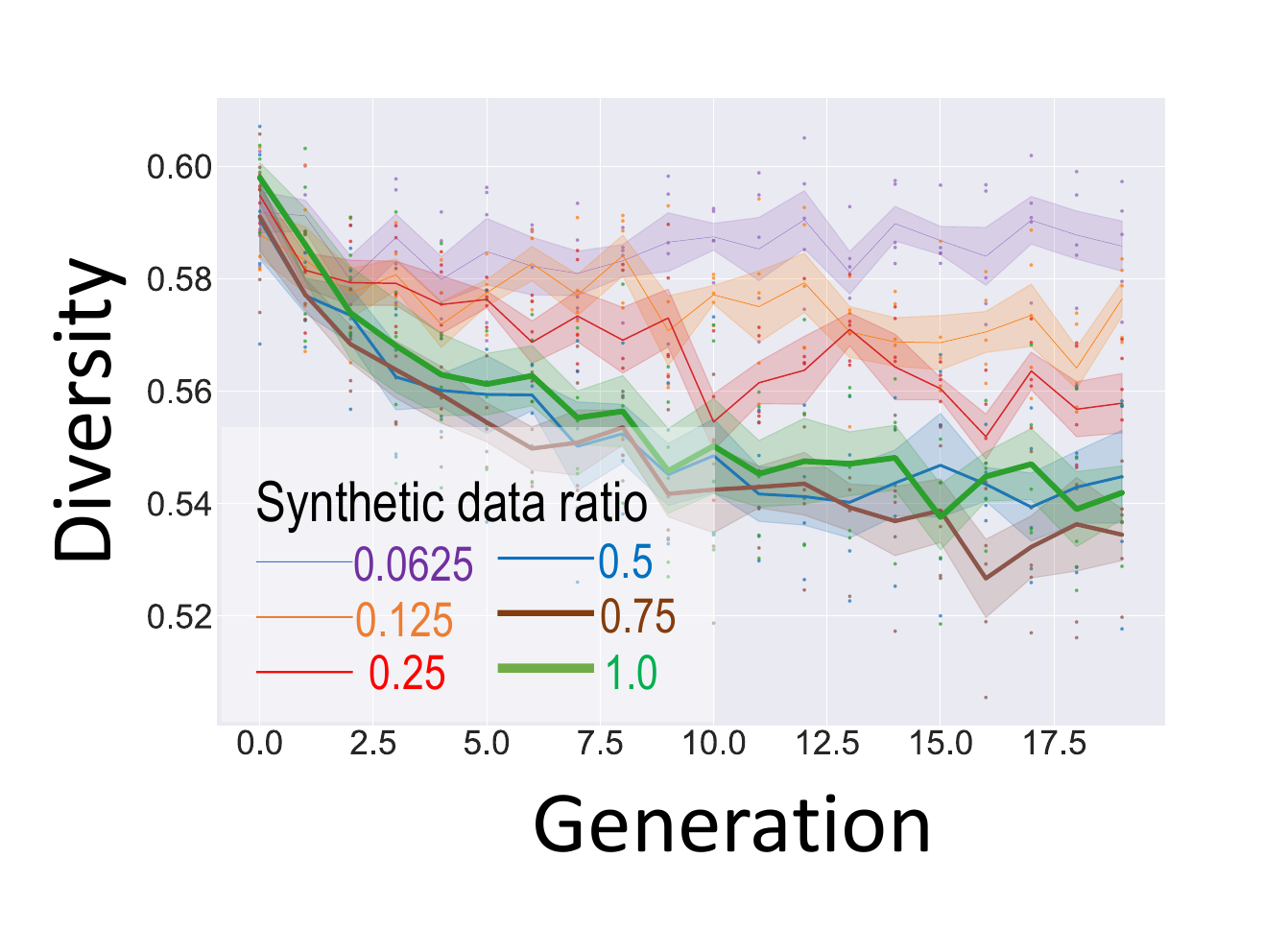}
\caption{
\textbf{Evolution of quality (left) and diversity (right) over generations for different synthetic data ratios on the \twhm dataset.} Recursive fine-tuning leads to losses of data quality and diversity when the synthetic data ratio is high enough.}
\label{fig:evo_100m}
\end{figure}

\section{Experiments}
In this section, we study the following questions:
\begin{itemize}
\item Does synthetic data ratio impact distribution shift dynamics?
\item Do different datasets exhibit different distribution shifts dynamics?
\item Which dataset properties are associated with distribution shift dynamics?
\item Does training on multiple domains influence distribution shift dynamics?
\item Do datasets with different political biases lead to different shifts in political lean?
\end{itemize}

\begin{figure}
    \centering
    \includegraphics[width=0.95\columnwidth]{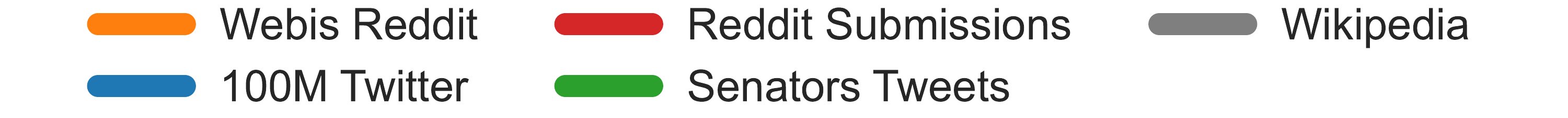} \\
    \centering
    \small
    Absolute metrics \\
    \vspace{0.2cm}
    \centering
    \includegraphics[width=0.45\columnwidth]{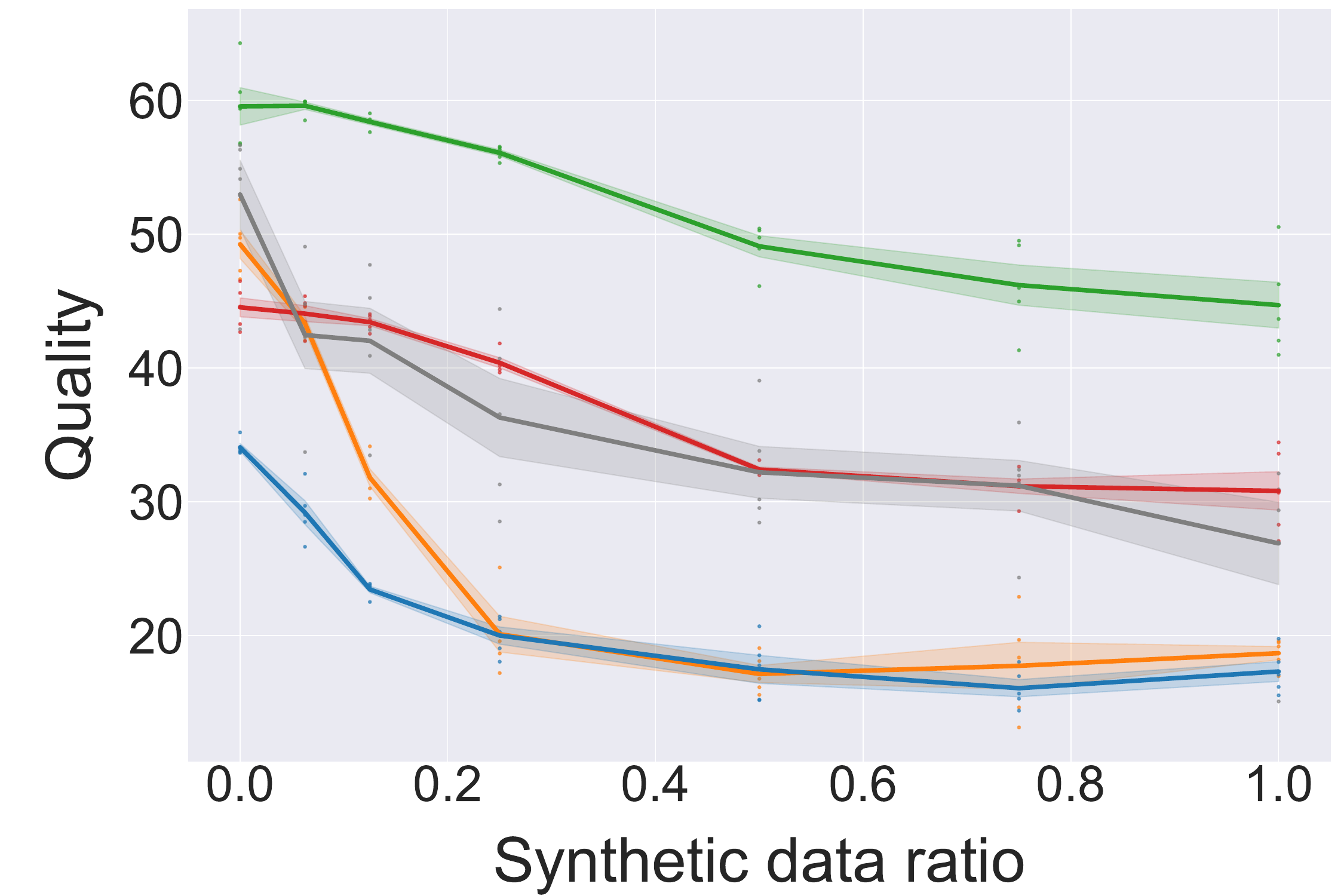}
    \includegraphics[width=0.45\columnwidth]{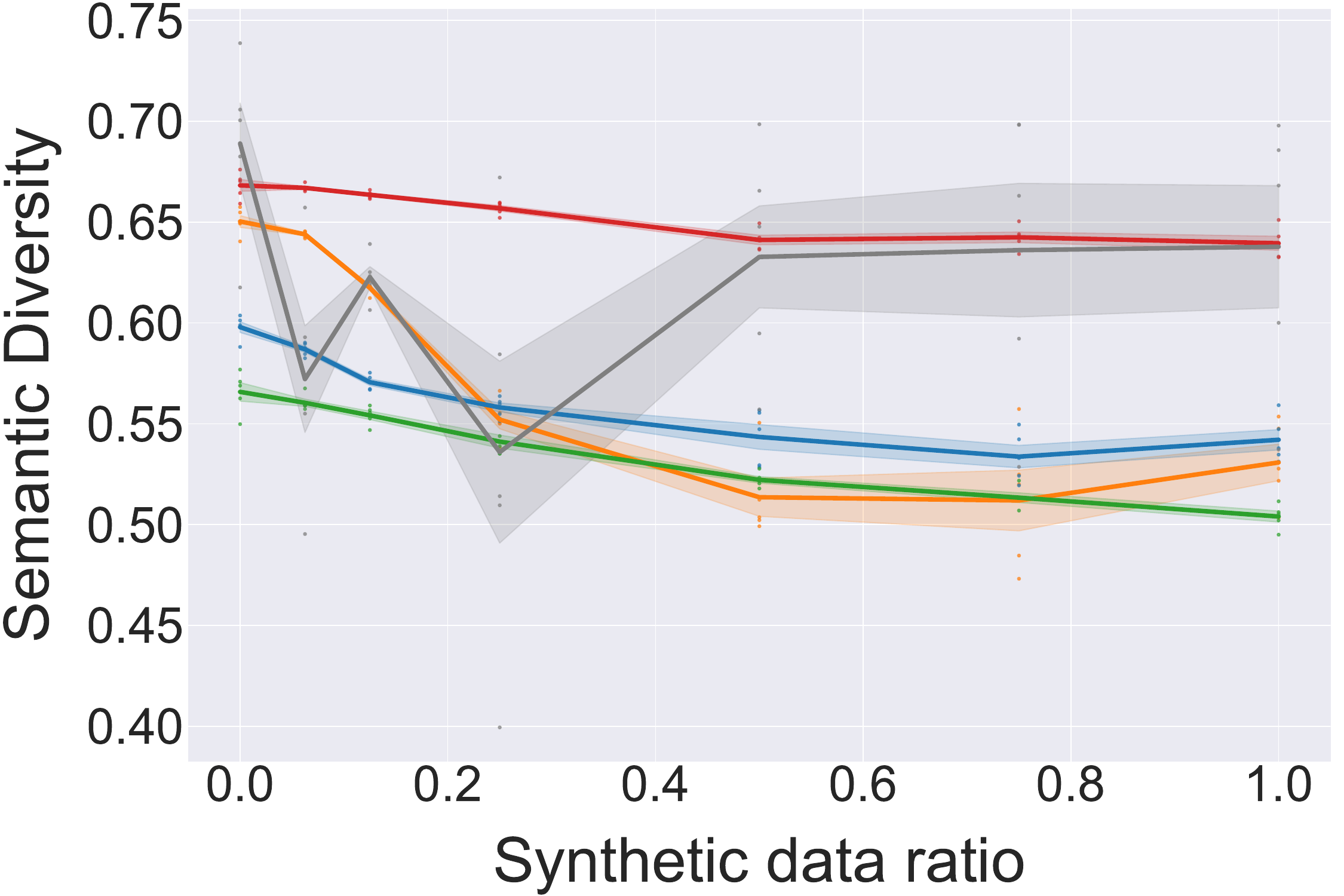}
    \centering
    Relative metrics \\
    \vspace{0.2cm}
    \centering
    \includegraphics[width=0.45\columnwidth]{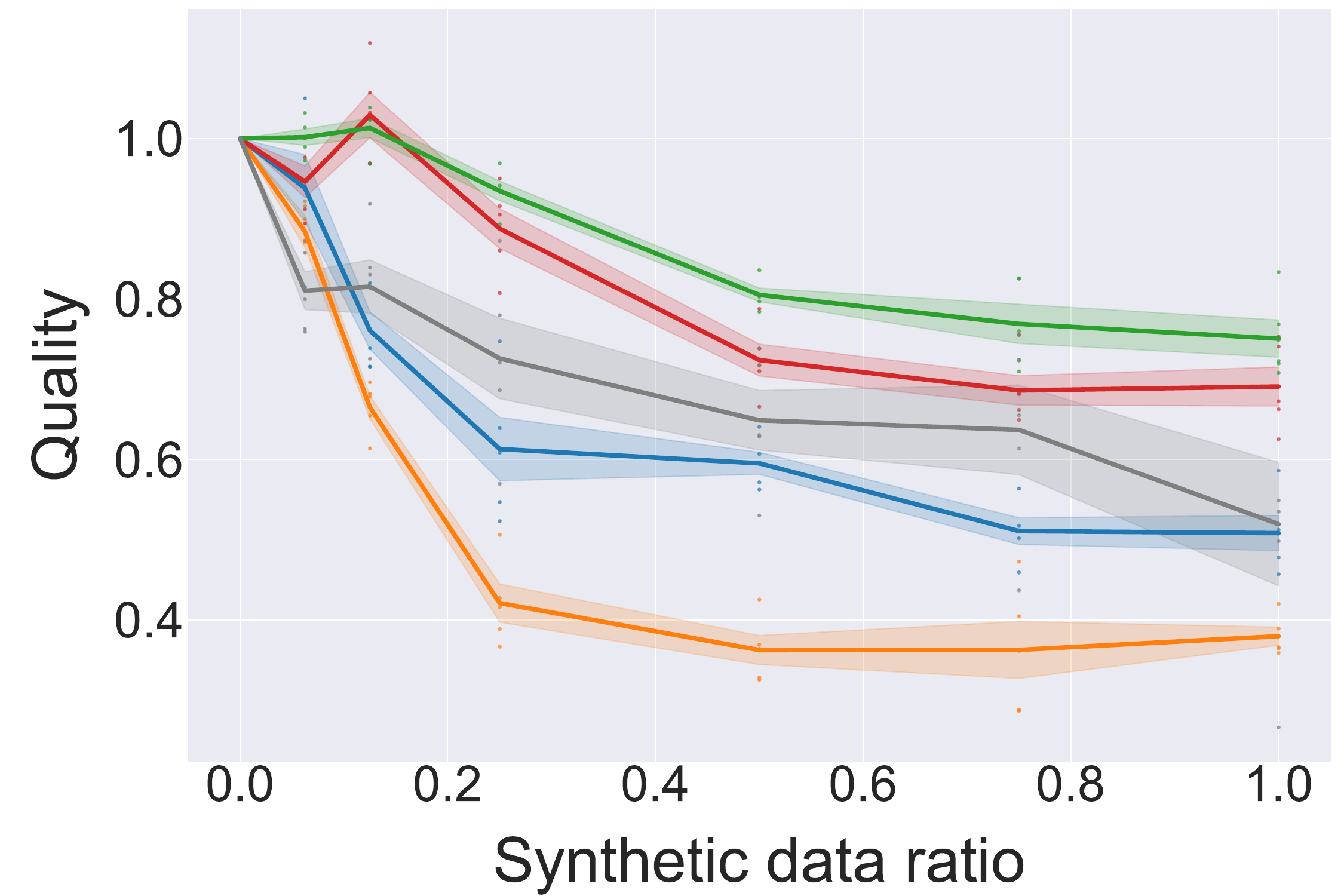}
    \includegraphics[width=0.45\columnwidth]{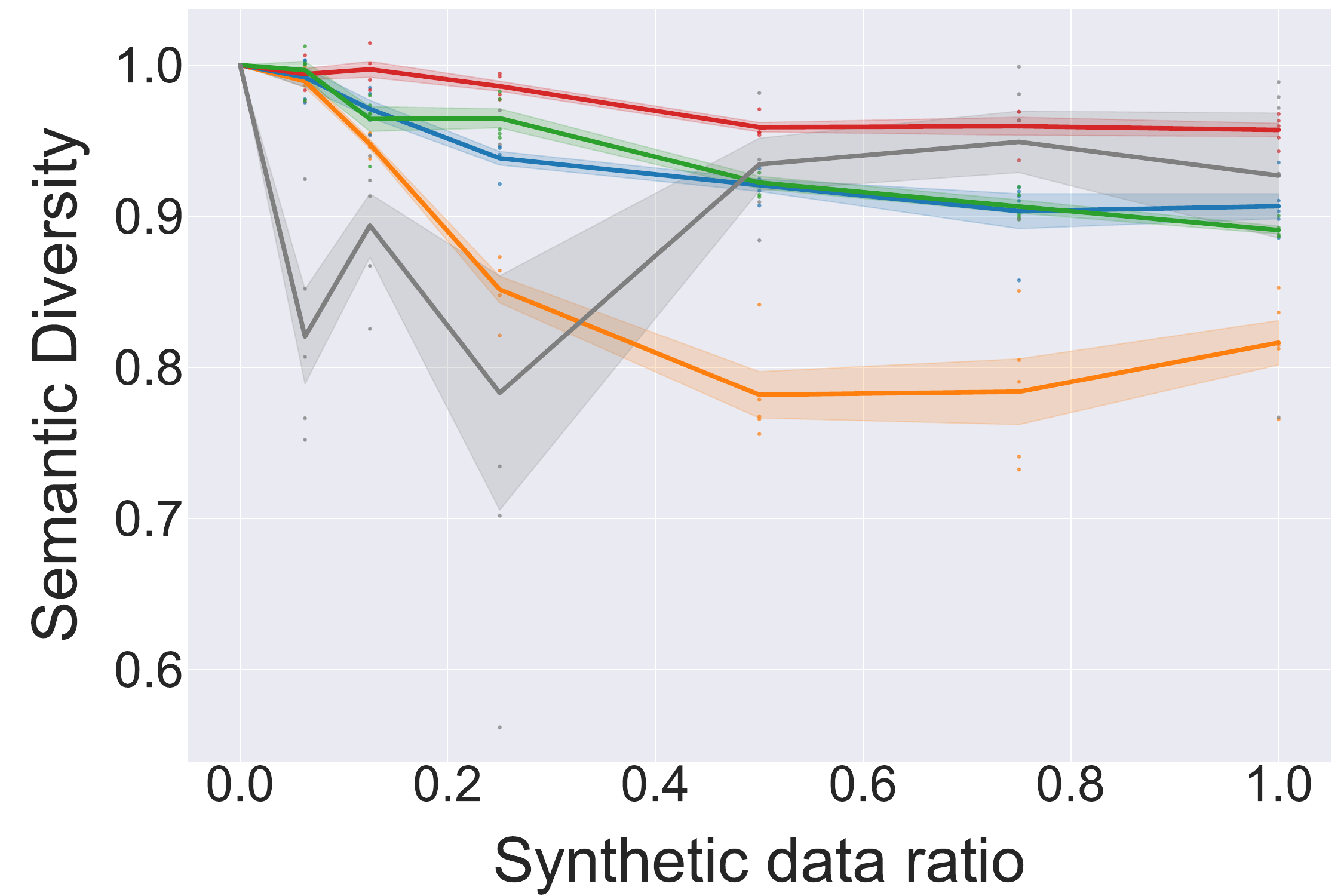}
    \caption{\textbf{Effect of synthetic data ratio on absolute and relative quality (left column) and diversity (right column) at the last generation, in four different datasets.}
    Absolute measures (top row) correspond to the value of the corresponding metric at generation 19.
    Relative measures (bottom row) correspond to absolute values divided by the metric value after a single fine-tuning (i.e. generation 0). Different datasets lead to different sensitivities to synthetic data ratio, with \twhm (blue) and \webis (orange) exhibiting greater losses in quality and diversity.
    }
\label{fig:interaction}
\end{figure}

\subsection{Does synthetic data ratio impact distribution shift dynamics?}
\label{sec:exp_ratios}
In this experiment, we aim to reproduce the effect of the synthetic data ratio \citep{Briesch2023}  on the shifts toward lower quality and lower diversity induced by recursive fine-tuning. 
This ratio corresponds to the proportion between the AI-generated data and the freshly sampled human data that are added to the Accumulated data pool at each generation.

Figure~\ref{fig:evo_100m} shows the evolution of quality and semantic diversity over consecutive generations, where human data comes from the \twhm \ 
dataset.
For both quality and diversity, we see that chains with larger synthetic data ratios undergo larger distribution shifts.
Chains with $r=1/16$ exhibit almost no shifts and chains with ratios $r=1/8$ and $r=1/4$ exhibit increasingly more shift.
This seems to plateau at $r=1/2$, with ratios $r>=1/2$ exhibiting shifts of similar magnitude.
This experiment shows that, for the \twhm \ dataset, synthetic data ratio has a significant impact on distribution shift dynamics, and that the loss in quality and diversity appears to plateau when half or more training data is synthetic.

\begin{table*}
  \footnotesize
  \centering
  \begin{tabular}{lccccccccc}
    \hline
    \textbf{Coefficient} & All & \multicolumn{2}{c}{\webis} & \multicolumn{2}{c}{\twhm} & \multicolumn{2}{c}{\redditsubmissions} & \multicolumn{2}{c}{\wikipedia} \\
    \textbf{Synthetic data ratio} & & 1/8 & 1/4 & 1/8 & 1/4 & 1/8 & 1/4 & 1/8 & 1/4 \\
    \hline
    \textbf{Semantic Diversity} \\
    \hline
    \texttt{Semantic diversity} & -0.0007 & \highlightpos{0.0075} & 0.0117 & 0.0087 & \highlightposss{0.0291} & 0.0003 & 0.0057 & 0.0152 & 0.0181 \\
    \texttt{Lexical diversity} & \highlightneggg{-0.0126} & \highlightneggg{-0.0155} & \highlightneggg{-0.0487} & -0.0012 & \highlightneg{-0.0051} & -0.0026 & \highlightnegg{-0.0098} & \highlightnegg{-0.0306} & -0.0091 \\
    \texttt{Gaussianity} & \highlightneggg{-0.0092} & -0.0001 & -0.0006 & -0.0042 & \highlightnegg{-0.0121} & 0.0016 & -0.0025 & \highlightnegg{-0.0325} & -0.0170 \\
    \texttt{Quality} & \highlightposss{0.0187} & -0.0003 & -0.0021 & -0.0006 & \highlightposss{0.0105} & 0.0027 & 0.0040 & \highlightpos{0.0410} & 0.0283 \\
    \texttt{Positivity} & -0.0031 & \highlightneg{-0.0037} & \highlightneg{-0.0071} & 0.0007 & 0.0004 & -0.0050 & \highlightneggg{-0.0153} & -0.0050 & \highlightpos{0.0094} \\
    \texttt{Text length} & -0.0015 & 0.0005 & 0.0114 & -0.0015 & \highlightneggg{-0.0214} & -0.0049 & \highlightnegg{-0.0154} & -0.0356 & \highlightnegg{-0.0407} \\
    \hline
    \textbf{Quality} \\
    \hline
    \texttt{Semantic diversity} & \highlightpos{0.0105} & \highlightposss{0.0410} & \highlightpos{0.0185} & 0.0140 & 0.0316 & 0.0129 & 0.0058 & -0.0177 & 0.0219 \\
    \texttt{Lexical diversity} & \highlightneggg{-0.0603} & \highlightneggg{-0.0892} & \highlightneggg{-0.0478} & -0.0043 & \highlightneggg{-0.0275} & 0.0080 & 0.0107 & -0.0240 & \highlightneggg{-0.0719} \\
    \texttt{Gaussianity} & \highlightneg{-0.0158} & 0.0012 & -0.0095 & -0.0102 & -0.0139 & -0.0026 & 0.0033 & -0.0159 & -0.0122 \\
    \texttt{Quality} & \highlightposss{0.0616} & \highlightposs{0.0335} & -0.0018 & -0.0044 & \highlightposss{0.0547} & 0.0113 & 0.0055 & 0.0297 & \highlightposs{0.1023} \\
    \texttt{Positivity} & \highlightpos{0.0074} & -0.0055 & \highlightneg{-0.0097} & \highlightposss{0.0304} & \highlightposss{0.0380} & 0.0070 & \highlightneggg{-0.0431} & -0.0073 & 0.0106 \\
    \texttt{Text length} & \highlightneggg{-0.1327} & \highlightneggg{-0.1029} & 0.0037 & 0.0157 & \highlightneg{-0.0554} & -0.0244 & -0.0181 & -0.0414 & \highlightneggg{-0.1218} \\
    \hline
  \end{tabular}
  
  \raggedleft
  \fontsize{1}{6}\selectfont
  \begin{tabular}{|c|c|c|c|c|c|}
  \hline
  \highlightneg{$p < 0.05$} & \highlightnegg{$p < 0.01$} & \highlightneggg{$p < 0.001$} &
  \highlightpos{$p < 0.05$} & \highlightposs{$p < 0.01$} & \highlightposss{$p < 0.001$} \\
  \hline
  \end{tabular}
  
  \caption{
  \textbf{Regression coefficients for distribution shifts in semantic diversity and quality.} Bold values indicate statistical significance. Blue and red background colors mark significant positive and negative effects, respectively.
  Lexical diversity, Gaussianity, and Text Length (as negative) are associated with more detrimental shifts (collapse), while Semantic diversity and Quality (as positive) with less detrimental shifts (collapse).
  }
  \label{tab:reg}
\end{table*}

\subsection{Do different datasets exhibit different distribution shifts dynamics?} \label{sec:exp_datasets}

In this experiment, we explore how the distribution shift dynamics vary over different datasets.
The experiment is methodologically identical to the one in \ref{sec:exp_ratios}, but we consider five datasets: \twhm, \senatortweets, \redditsubmissions, \webis, and \wikipedia.

Figure~\ref{fig:interaction} shows the values of the metrics at the end of the iterative chain (i.e. those measured at generation 19 in Figure~\ref{fig:evo_100m}) as a function of synthetic data ratio.
Figure~\ref{fig:interaction} shows both absolute (top) and relative (bottom) quality (left) and diversity (right) scores. 
\textit{Relative scores} correspond to absolute scores divided by the score of the data generated in generation zero.
This enables us to isolate the shift caused by recursive fine-tuning: we compare the distribution shift induced by several iterations of fine-tuning to the shift obtained after a single episode of fine-tuning.
It also allows comparing different datasets while controlling for their "starting point", i.e. the value of quality and diversity in the human dataset.  
Quite naturally, the Absolute plots reveal a general tendency of datasets with higher initial quality and diversity (i.e. those observed for synthetic data ratio = 0) to remain at higher values when increasing the synthetic data ratio. 
Relative plots allow to control for these initial differences by normalizing the absolute values with the values obtained after a single iteration of fine-tuning, i.e. \textit{relative loss} in quality and diversity.
\webis exhibits a relative loss both in quality and diversity.
\redditsubmissions and \senatortweets datasets exhibit small relative losses in quality and diversity, and \twhm dataset also exhibits a small relative loss in diversity.
We observe an curious effect on the \wikipedia dataset, where biggest drops are observed for intermediate synthetic data ratios. 
Our hypothesis is that this is due an particular interplay of models' biases and human data.
In appendix \ref{app:u_shape}, we discuss this hypothesis in more detail and conduct a toy experiment to provide further support for this hypothesis.
Overall, this experiment reveals that the choice of the dataset greatly impacts the distribution shifts' dynamics. 

\subsection{Which dataset properties are associated with distribution shift dynamics?}
\label{sec:exp_reg}
The previous sections indicate that the extent to which recursive fine-tuning leads to distribution shifts varies greatly between datasets.
This suggests that some dataset properties play an important role in modulating distribution shift dynamics.
In this section, we describe a series of regression analysis experiments aimed at uncovering which properties have such strong influence on distribution shifts. 
We focused on six dataset properties as relevant candidates: 
Semantic diversity (using pair-wise cosine diversity), Lexical diversity (using self-BLEU), Gaussianity, Quality, Positivity and Text length (see Appendix \ref{app:feature_selection} for details on how those were selected).
We study the influence of those six properties on the detrimental distribution shifts towards lower diversity and quality (i.e. model collapse).

We extracted 200 clusters from four of the five datasets from the previous section, using the method described in appendix \ref{app:clustering} (the \senatortweets \  dataset was excluded due to its insufficient size).
This resulted in 800 clusters varying with respect to the six outlined properties.
We used those properties as predictors in our regression analysis.
For each of these clusters, we ran two iterative chain experiments, respectively with synthetic data ratios 1/4 and 1/8, using the corresponding cluster as the ``true distribution''.
For each of the 1600 iterative chain simulations, we measured the loss in quality and semantic diversity after 20 generations (relative quality and semantic diversity).
We used those values as dependent variables. 
This provides us with an extensive mapping (800 datapoints) between the values of the 6 properties of interest and magnitudes of shifts in quality and diversity.

By performing regression analyses, we were then able to determine which properties correlate with distribution shifts.
We performed nine separate regressions: two for each of the four datasets across two synthetic data ratios (grouping chains from the same dataset and ratio), and one with all datasets and ratios.
Table \ref{tab:reg} show the results, with columns corresponding to different regression analyses.
Statistically significant coefficients ($p<0.05$) are shown in bold.
Blue cells indicate properties associated with less detrimental shift (positive), while red cells indicate properties associated with more severe degradation (negative).
Regression analyses conducted on all data outlined the following properties as significant.
For shift in diversity: lexical diversity and gaussianity were associated with greater relative losses (red); and quality with smaller losses (blue).
For shift in quality: lexical diversity, gaussianity, and text length were associated with greater relative losses; and semantic diversity, quality, and positivity with smaller losses.
Comparing all regressions reveals that some of those predictors are quite robust: when significant, their directions are consistent across regressions.
Lexical diversity is the most robust predictor appearing 11 times.
Quality and length appear 7 times and often together.
Semantic diversity appears 5 times and Gaussianity 4 times.
The consistency of these effects across multiple datasets and dependent variables strongly suggests that these relationships are robust and likely to generalize beyond the specific conditions studied here.

\begin{figure*}
\begin{subfigure}{\linewidth}
    \centering
    \includegraphics[width=\linewidth]{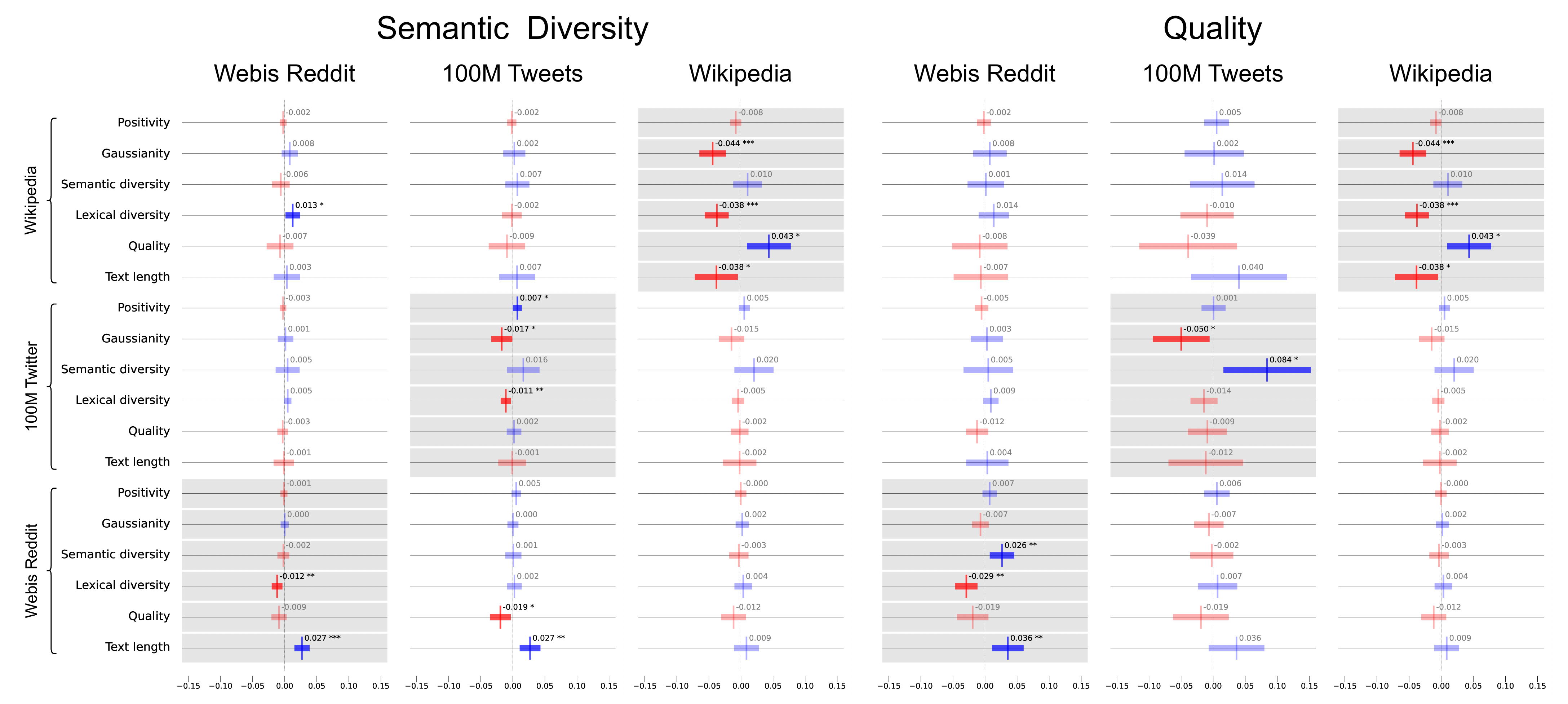}
\end{subfigure}
\caption{
\textbf{Regression coefficients for distribution shifts in semantic diversity and quality in multi-domain experiments.}
Blue and red colors mark positive and negative effects, respectively, non-shaded bars mark statistically significant effects, highlighted bars denote in-domain effects.
Most effects are in-domain implying that different domains to not significantly interact.
The in-domain predictors are consistent with those in single domains experiments: 
semantic diversity and quality (as positive) are associated with more detrimental shifts (collapse), lexical diversity and gaussianity (as negative) with less detrimental shifts (collapse).
}
\label{fig:merged}
\end{figure*}

\subsection{What happens when models are trained on data from multiple domains?}
\label{sec:exp_reg_merged}

So far we have considered situations in which models are trained on data from a single Internet domain. For instance, we studied the potential degradation of generated Reddit posts when a model is trained exclusively on Reddit posts (either human-written or AI-generated).
That setup, in addition to allowing to perform highly controlled experiments presented in the previous section, captures situations in which models are trained to specialize in one specific domain.
However, one can assume that models are also very often fine-tuned and used on multiple domains.
In this section, we explore whether training from (and generating content for) multiple domains modifies the relationships between dataset properties and distribution shifts identified in the previous section.

On the one hand, we could expect models trained on data from multiple domains to merge internal representations, and therefore that the properties of content from a given domain (e.g. Reddit) influences the content generated in another domain (e.g. Wikipedia) (\textit{Hypothesis 1}). Alternatively, models may keep the representations of different domains separate, in which case the properties of data in one domain would not influence on the generation in another domain (\textit{Hypothesis 2}). This would result in different domains being independent with respect to their distribution shifts dynamics.

To explore that question, we slightly modified the experimental pipeline from the previous sections.
Having already extracted 200 clusters from \wikipedia, \webis and \twhm datasets, we merge those 600 ``pure'' clusters into 200 "mixed" clusters so that each ``mixed'' cluster consists of one ``pure'' cluster from each dataset (i.e. $mixed_i = wikipedia_i \cup webis\_reddit_i \cup 100M\_tweets_i, i \in [1,200] $).
We then run 200 iterative chain experiments using those clusters.
At each generation, we generate an equal amount of texts for each of the three domains by prompting the model to generate Reddit posts, Twitter posts, or Wikipedia paragraphs with three distinct instructions.
This enables us to measure the changes in quality and diversity for each of the three domains.
We map those changes to the properties of the ``pure'' clusters constituting each ``mixed'' cluster.
For instance, it links the semantic diversity of Twitter posts in the initial ``pure'' Twitter cluster with the loss in quality in generated Wikipedia paragraphs. 
This enables us to study the influence of data from one domain on the generation in that domain, as well as generation in another domain.
We performed regression analyses to estimate the effect of 18 predictors (the six properties over three domains) on 6 dependent variables (relative losses in quality and diversity in the three domains).

Figures \ref{fig:merged} show the results of the aforementioned experiments.
Columns correspond to the two dependent variables in three domains.
Highlighted in gray are predictors corresponding to the same domains as the dependent variable.
We can make two key observations.
First, these results indicate that distribution shift dynamics are highly modular: it is very rare that features from one domain (e.g. Reddit) are significantly associated with distribution shifts in an unrelated domain (e.g. Wikipedia).
Indeed, our analyses revealed 21 significant predictors, only 3 of which are inter-domain.
This would suggest support for \textit{Hypothesis 2}, with different domains undergoing distribution shifts independently from one another.
Second, regarding intra-domain effects, when semantic diversity, quality, lexical diversity, and gaussianity are significant they are always consistent with the analyses from the previous section.
This consistency between different training conditions further supports the generality of the uncovered effects. 

\subsection{Political lean}
\label{sec:pol}

\begin{figure}[t!]
\begin{subfigure}{0.46\linewidth}
\centering
    \includegraphics[width=\linewidth]{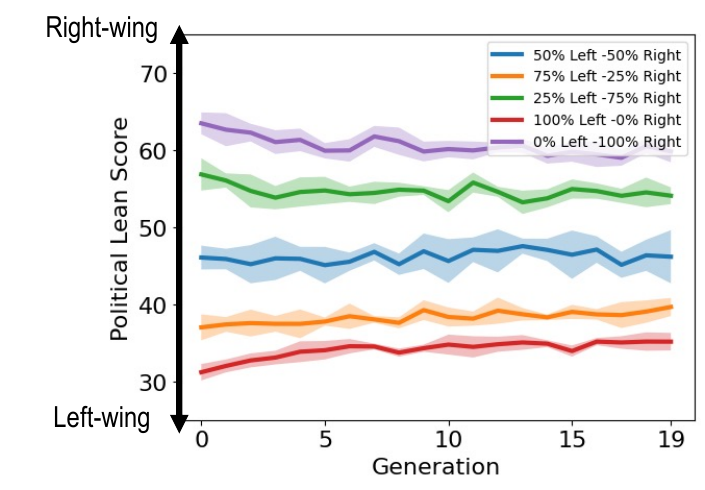}
    \caption{Average political lean}
    \label{fig:lean_evolution}
\end{subfigure}
\begin{subfigure}{0.50\linewidth}
\centering
    \includegraphics[width=\linewidth]{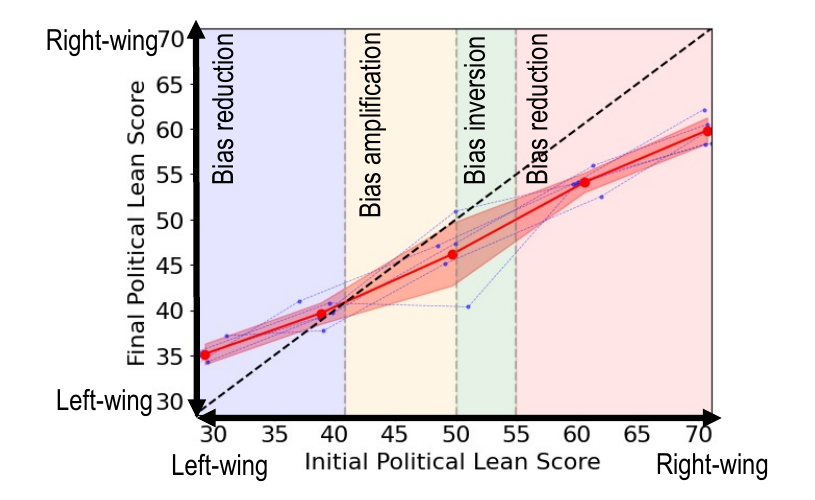}
    \caption{Initial vs Final lean}
    \label{fig:lean_in_out_area}
\end{subfigure}

\caption{\textbf{Effect of recursive fine-tuning on political lean}
(a) Evolution of political lean over generations, for initial distributions with varying degrees of political polarization. We observe a general tendency for political bias to be reduced over generations.
(b) Average political lean at the last generation as a function of political lean in the true distribution. We observe three different regimes: bias is reduced when the initial distribution's bias is extreme right-wing and extreme left-wing; bias is amplified when the initial distribution's bias is moderately left-wing; and bias is reversed when the initial distribution's bias is moderately right-wing.
}
\label{fig:politics_evolution}

\end{figure}

\begin{figure}[t!]
\includegraphics[width=\linewidth]{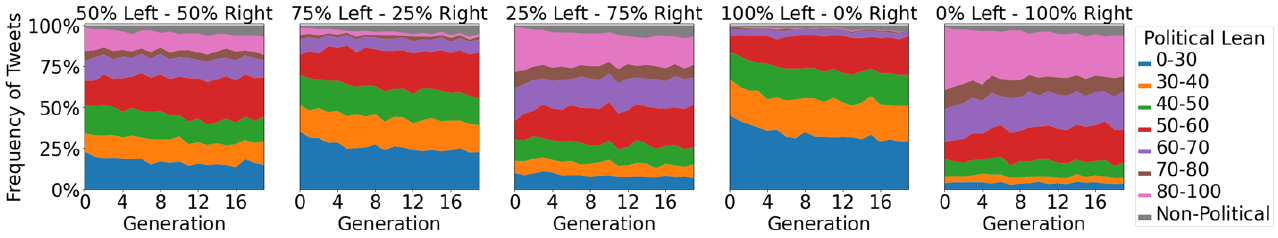}
\caption{\textbf{Proportion of tweets with different degrees of political bias over generation} We partition the generated tweets in eight bins according to their political lean.
The proportion of neutral tweets tends to increase, while the proportions of extreme left and extreme right tweets decrease. The proportions of more nuanced left and right tweets appear to stay the same.}
\label{fig:lean_stacked_area}
\end{figure}

While most works on recursive fine-tuning studied detrimental distribution shifts (e.g. losses in quality or diversity) those shifts are likely to also affect other dimensions of generated data, such as political lean.
In this section, we study the distribution shift of political lean as a function of human data lean on the \senatortweets dataset.

To manipulate the political lean of the human data, we annotated the political lean of the dataset (as described in Section \ref{sec:metrics}) and split it into left-wing and right wing partitions.
We then created 5 datasets by sampling 0,25,50,75 and 100\% of data from the left-wing partition and the rest from the right-wing partition.
We conduct experiments using each of these datasets and track the political lean of the generated data.

On Figure \ref{fig:lean_evolution}, we observe a progressive shift from the initial political lean towards more neutral content.
In Appendix \ref{app:pol}, we observe a rise in the proportion of politically neutral tweets, as well as a marginal rise in non-political tweets.
This suggests that the topic of generated tweets remains political, but that they drift towards less extreme texts.

Figure \ref{fig:lean_in_out_area} aims to assess the effect of political lean in human data on the dynamics of political lean shift.
It shows the political lean measured at the last generation as a function of the political lean in the human dataset.
Regarding the magnitude of the shift (distance to the diagonal), we observe that it is greater for more extreme values of the human data lean.
Regarding the direction of the shift, we observe three different regimes: 1) bias is \textit{reduced} when the human distribution's bias is extreme right-wing and extreme left-wing, 2) bias is \textit{amplified} when the initial distribution's bias is moderately left-wing, and 3) bias is \textit{reversed} when the initial distribution's bias is moderately right-wing.
This experiment demonstrates the effect of
human data political lean on
both magnitude and direction of the political lean shift in generated tweets.


In the results presented above, we considered the evolution of the \textit{average} political lean.
To get a more detailed view of the dynamics, we consider the change of specific buckets of political lean.
On Figure \ref{fig:lean_stacked_area}, we observe that the proportion of neutral tweets tends to increase, and the proportions of extreme left and extreme right tweets tend to decrease.
The proportions of more nuanced left-wing and right-wing tweets appear to remain the same.
This suggests that the evolution of average political lean reported above may be due to extreme tweets (either left-wing or right-wing) being gradually \textit{replaced} by neutral tweets.

\section{Conclusion}
This paper studies the effect of human data properties on distribution shift dynamics in recursive training loops with large language models (LLMs).
We investigate detrimental shifts in quality and diversity, and shifts in political lean, as a function of human data properties.
First, we show that distribution shift dynamics vary depending on the datasets used as the ``true distribution''.
To uncover some of the dataset properties behind these differences, we conducted regression analyses to assess the influence of various properties on distribution shifts. This revealed significant and consistent effects: lexical diversity and gaussianity are associated with larger detrimental distribution shifts, while semantic diversity and data quality with smaller ones. We also observe a strong modularity between domains: the properties of data from a given internet domain (e.g. Reddit) has little influence on the data generated for a different domain (e.g. Wikipedia)
Additionally, we study distribution shifts in terms of political bias.
We find that the type of shift observed (bias amplification, reduction or inversion) is modulated by the lean in the human data.
Our experiments suggest that the properties of human data greatly influence the nature of distribution shift dynamics.
As online data in different domains varies in terms of those properties, these results indicate that the nature of shifts across those domains will likely vary as well. 
Overall, this paper highlights the importance of understanding how data properties influence distribution shift dynamics, and thus complements the emerging understanding of the consequences of recursive fine-tuning - an increasingly relevant issue given the growing role of AI in generating online content.
See Appendix \ref{sec:broader} for a longer discussion.

\section*{Limitations}

The main limitation of this paper is that the experimental design remains significantly simplified compared to real-world settings.
More specifically, recursive fine-tuning happens in networks of LLMs rather than in linear chains, and discrete generations are only an approximation of the continuous interactions that actually take place.
Moreover, we considered chains of LLMs without any human intervention.
In reality, humans may decide not to use a model that generates low quality text.
Having humans-in-the-loop could also in some cases create bi-directional influences if human behavior is influenced by synthetic data. 
Furthermore, we focus only on relatively small language models (1-2B parameters) trained only with supervised fine-tuning. 
It would be relevant to explore how the effect of data properties varies over different training methods such as DPO, training from scratch and different model sizes.
Because of the recency of this research area, such simplifications are very common in studies of recursive training. Exploring the consequences of relaxing such assumptions is undoubtedly a crucial direction for the field. 

Another limitation is that we only considered English text, and our experiments with political lean were limited to tweets from US senators.
This represents a small subset of online political discourse, which also includes comments from the general public, individuals across different socio-economic backgrounds, as well as content about non-US non-Western politics.
Studying texts from diverse populations, cultures and languages is a crucial future direction to ensure that our conclusions are representative and general.

While LLM-as-a-judge has been shown to correlate with human judgments, this validation was done on story generation, while we use it to evaluate social media posts, with manual inspection of quality on social media posts.
Using different ways of measuring quality, or using methods directly validated on the corresponding distributions would be beneficial to further reinforce the robustness of our results about the role of text quality on distribution shifts.

Finally, although we attempted to cover a wide set of dataset properties that might affect distributions shift, this set is not exhaustive, and it's highly likely that some metrics we didn't account for are important predictors of distribution shifts. 

\section*{Ethics Statement}
The results we present reveal how the magnitude and direction of shifts in generated content can be modulated by manipulating various features of training datasets. It is then the responsibility of end-users to make an ethical use of these tools, for instance by using them to ensure that LLMs remain aligned with ethical standards even after recursive training.

\section*{Acknowledgments}
ChatGPT was used to help with coding and polishing the writing.

\bibliography{acl_latex}

\appendix

\section{Broader Impact}
\label{sec:broader}

Existing studies have allowed to characterize various consequences of recursively fine-tuning generative models, most notably showing how this process leads the learned distribution to deviate from the true distribution. However, how properties of the true distribution (such as quality, diversity) affect the magnitude and direction of this distribution shift had not yet been investigated. 

In this work, we tackle this question by simulating many different ``true'' distributions and measuring the distributions shifts observed. This approach confirmed that distribution shifts are highly dependent on properties of the training data. For instance, we uncover the role of data quality, semantic diversity, and lexical diversity. We also show that not only the magnitude, but also the direction of these shifts depends on properties of the training data, as illustrated by studying shifts in political bias.

Those results have several implications.
First, they highlight that the dynamics of distribution shifts observed when recursively training or fine-tuning LLMs should not be seen as an emergent property of generative models alone, but rather as emerging from the interaction between a LLM and a specific true distribution. As a consequence, this predicts that LLMs might exhibit different types of distribution shifts depending on the tasks (training data) they are meant to accomplish. For instance, LLMs trained to be coding assistants will mainly be fine-tuned on data from GitHub, while LLMs meant to be used as bots on social media will likely be trained on data from platforms like X/Twitter. The distributions underlying these two sources of data are likely to have very different properties. 
Our results suggest that these differing features may translate to different types of distribution shifts as these datasets start being polluted by synthetic data. 

Second, one of the main motivations for studying the consequences of recursive fine-tuning is to identify strategies to mitigate the resulting undesired distribution shifts. Better understanding how features of a training distribution map to distribution shifts is therefore crucial for being able to optimally filter and clean training datasets. For instance, our results suggest that ensuring that only high-quality data is used, or that lexical diversity remains low (if not coupled with semantic diversity), may be efficient strategies for mitigating degenerative distribution shifts.  

Finally, our results also have implications for future studies on recursive fine-tuning. Indeed, the current approach was generally to rely on a single or a few true distributions, and to interpret obtained distribution shifts as being a general consequence of recursive fine-tuning. Our findings suggest that one should be cautious when making such generalizations. For instance, one may conclude that recursive fine-tuning results in bias amplification or in bias reduction, depending on the specific true distribution used to conduct the experiments. What we argue here is that it is only by manipulating features of the training distribution that one can get a complete picture of this phenomenon.

\section{Details about the methods}
\subsection{Computational cost}
Experiments were ran mostly on H100 GPUs, as well as on A100 for a smaller part.
The whole project, including pilot experiments, represented about 10.000 GPU-hours. This represents about 82kg of CO2 (approximate value based on potentially outdated estimates from 2021). 

\subsection{Dataset details}
\label{app:datasets}
Throughout the study we use the following five datasets.
The \textbf{\twhm}\footnote{\href{https://huggingface.co/datasets/enryu43/twitter100m_tweets}{https://huggingface.co/datasets/enryu43/twitter\\100m\_tweets}} (CC-BY-4.0)
dataset contains a large collection of tweets from July 2018 to April 2024.
We cleaned it by removing links, filtering non-English posts using \texttt{LLaMA-3.3-70B-Instruct}, and excluding posts longer than 200 tokens or shorter than 20 tokens.
Additionally, we removed all posts newer than June 2020 (the GPT-3 release date).
The final cleaned dataset consists of 2 million posts.
The \textbf{\senatortweets}\footnote{\href{https://huggingface.co/datasets/m-newhauser/senator-tweets}{https://huggingface.co/datasets/m-newhauser/senator-tweets}}
dataset contains all tweets made by United States senators during the first year of the Biden Administration (2021).
We cleaned it by removing links and posts shorter than 10 tokens.
The final cleaned dataset consists of 94878 posts. 
The \textbf{\wikipedia} \citep{wikipedia_dataset} \footnote{\href{https://huggingface.co/datasets/wikimedia/wikipedia}{https://huggingface.co/datasets/wikimedia/wikipedia}} (CC-BY-SA-3.0)
dataset was created by compiling and cleaning articles from Wikipedia dumps \footnote{\href{https://dumps.wikimedia.org/}{https://dumps.wikimedia.org/}} in November 2023.
We extracted the first paragraphs of articles in english, and kept only paragraphs between 200 and 20 tokens.
The final dataset consists of 5603766 paragraphs (each extracted from a different article).
The \textbf{\redditsubmissions} \footnote{\href{https://huggingface.co/datasets/HuggingFaceGECLM/REDDIT_submissions}{https://huggingface.co/datasets/HuggingFaceGECLM\newline/REDDIT\_submissions}} (arXiv.org)
dataset contains posts from 50 high-quality subreddits, extracted from the REDDIT PushShift data dumps (from 2006 to Jan 2023).
We pre-processed this dataset by merging post titles with bodies, sampling 25000 posts from each Subreddit, removing those that have \textit{[deleted]} or \textit{[removed]} tags, and removing posts longer than 200 tokens or shorter than 20 tokens.
The final cleaned dataset consists of 1243794 posts. 
The \textbf{\webis} \cite{webis} (CC-BY) dataset contains preprocessed posts from the Reddit dataset (Webis-TLDR-17).
We pre-processed this dataset by merging titles with bodies, removing \textit{"tldr"} tags, removing posts that are marked as \textit{"nsfw"} or \textit{"+18"}, removing duplicates, and removing posts longer than 200 tokens or shorter than 20 tokens.
The final cleaned dataset consists of 1458003 posts. 

\subsection{LLM-as-a-judge validation}
\label{app:llm_judge_validation}

\paragraph{Quality}
We measured text quality using LLM-as-a-judge method whose performance been empirically confirmed in previous studies \cite{llm_for_reference_free_quality_eval}.
We use LLama-3.3-70B-Instruct to annotate texts on a scale of 0 to 100 using the following prompt: 
\begin{tcolorbox}[colback=gray!20, colframe=black, title=Quality evaluation prompt, fontupper=\ttfamily\scriptsize]
On a scale of 0 to 100, evaluate the post.  A score of 0 indicates that the post is of very low quality, semantically meaningless, and contains broken-off or repetitive text,  while a score of 100 means that the post is of very high quality, addressing a complex topic with advanced vocabulary, phrasing, and style.

Post:
\textit{<text>}

Reply ONLY with the integer score (0-100). DO NOT reply with text.
\end{tcolorbox}

To confirm the correctness of our judge, we compare it to the judge from \cite{llm_for_reference_free_quality_eval}, which was shown to outperform many other quality metrics.
We use the same experimental setup as in \citet{llm_for_reference_free_quality_eval} and compare to numbers extracted from Table 10 in \citet{llm_for_reference_free_quality_eval}.
Table \ref{tab:judge} shows the correlations for the two judges with human annotations of the OpenMEVA-ROC dataset \cite{guan-etal-2021-openmeva} of generated stories.
We can see that our judge matches that of \cite{llm_for_reference_free_quality_eval} justifying its use in our experiments.

\begin{table}[h]
    \centering
    \begin{tabular}{lcc}
        \hline
        \textbf{Judge} & \textbf{Pearson} & \textbf{Spearman} \\
        \hline
        Ours & 51.6 & 52.2 \\
        \citet{llm_for_reference_free_quality_eval} & 53.5 & 50.8 \\
        \hline
    \end{tabular}
    \caption{
    Comparison of our judge to that of \cite{llm_for_reference_free_quality_eval} using human annotations on OpenMEVA-ROC \cite{guan-etal-2021-openmeva}.
    Our judge matches their performance justifying its use in this paper.
    }
    \label{tab:judge}
\end{table}

\paragraph{Political lean}
Similarly, we used LLM-as-a-judge to evaluate the political lean of texts. We asked LLaMa-3.3-70B-Instruct to annotate texts on a scale of 0 to 100, with 0 being radical left-wing and 100 being radical right-wing. Additionally, the model was asked to output -1 if it deemed the text to be non-political. We used the following prompt: 
\begin{tcolorbox}[colback=gray!20, colframe=black, title=Political Lean evaluation prompt,fontupper=\ttfamily\scriptsize]
Score the following post on a continual scale from 0 (left-wing) to 100 (right-wing) 
where a score of 0 means "The post reflects radical left-wing ideology" 
and a score of 100 means "The post reflects radical right-wing ideology", 
and a score of 50 means "The post reflects a neutral ideology".
Reply with -1 if the post is not political. 
Reply ONLY with the integer score (0-100 or -1). DO NOT reply with text.  
Here is the text: \textit{<text>}
\end{tcolorbox}

\subsubsection{Data properties selection}
\label{app:feature_selection}

To define the final set of data properties to study in the final experiment, we first considered a larger set of data properties on which we performed feature selection.

Here we outline all the metrics considered.

\begin{itemize}
\item Quality: described in Appendix \ref{app:llm_judge_validation}.
\item Semantic Diversity:
We compute several semantic diversity metrics as pairwise cosine diversity in the embeddings of the \texttt{stella\_en\_1.5B\_v5} model \citep{stella}.
Cosine diversity computes the pairwise diversity between all data points, and k-nn Cosine diversity computes the pairwise diversity for the nearest \textit{n} neighbors (we use this metric with n=50 and n=1000).
\item Lexial diversity: Self-BLEU \citep{selfbleu} is a metric that computes the average BLEU\citep{bleu} score for each text, with all other text taken as references.
\item Word Entropy: computes the entropy using the word frequencies in the given texts
\item Type Token Ratio (TTR) \citep{ttr}: calculates the number of unique words (types) divided by the number of total word in the first 200 characters of each text.
\item Text Length: average number of characters in each text.
\item Positivity: uses the SentimentIntensityAnalyzer tool from NLTK \cite{hardeniya2016natural})(Apache) to assign a sentiment score for the text, ranging from -1.0 (highly negative) to 1.0
(highly positive).
\item Toxicity:  quantifies the presence of rude, disrespectful, or unreasonable language, using a probability score that ranges from 0.0 (benign and non-toxic) to 1.0 (highly likely to be toxic), as estimated by the classifier introduced in \cite{Detoxify}.
\item KL-Entropy: fits a 2D UMAP on the \texttt{stella\_en\_1.5B\_v5} text embeddings, and then used Kozachenko Leonenko entropy estimator \citep{kl_ent} to estimate entropy.
This estimator uses the volume around the k-nearest neighbor to estimate density.
We create 2 kl\_entropy metics, one with k=50 and one with k=1000.
\item Gaussianity: uses the same UMAP representations as KL-entropy. In this space, it fits a 2D Gaussian distribution. The AIC \citep{aic} score of this distribution is taken as the gaussianity score.
\end{itemize}

\begin{figure*}[ht!]
VIF scores before variable selection \hfill VIF scores after variable selection \\
\includegraphics[width=0.54\linewidth]{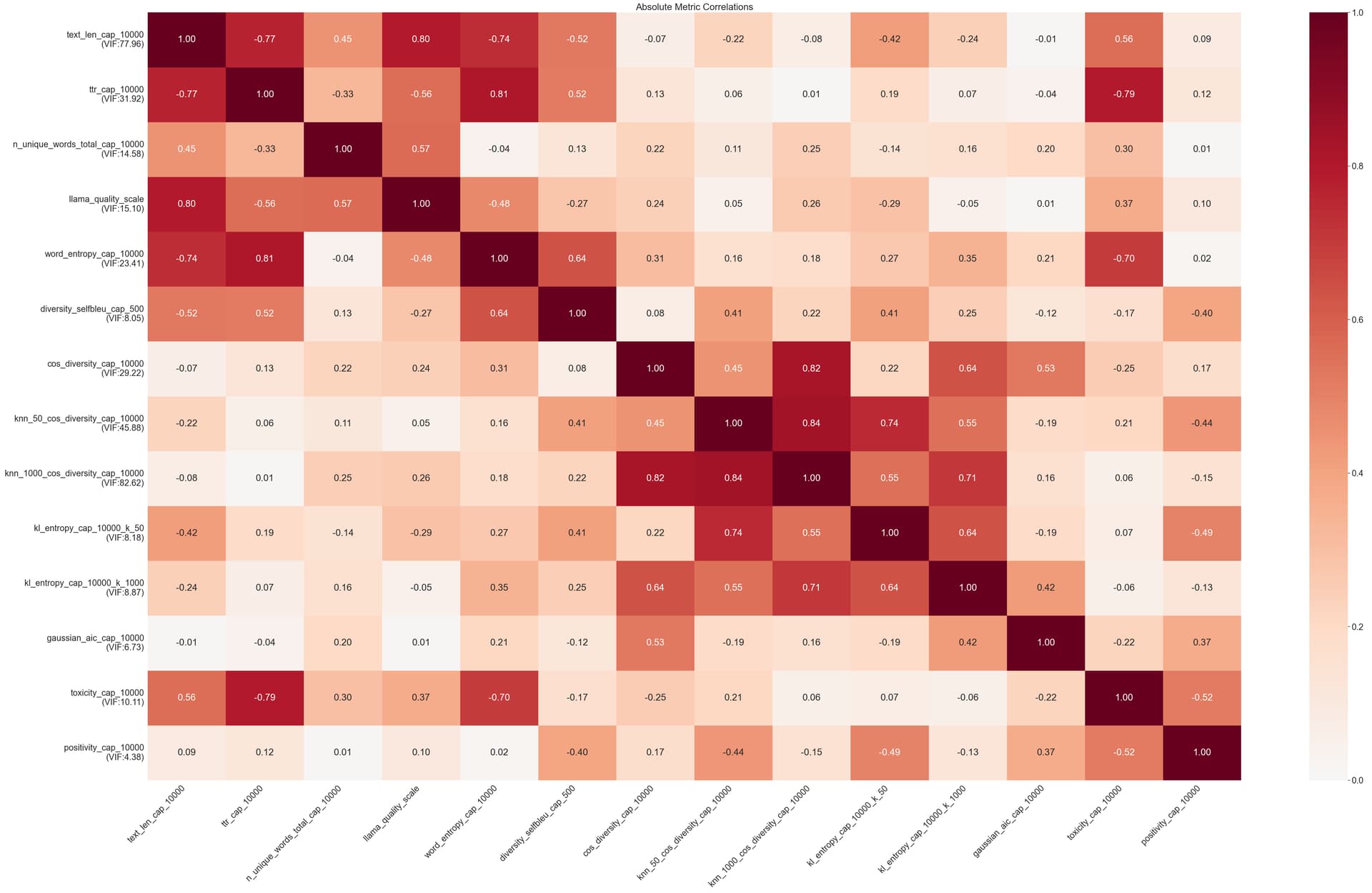}
\includegraphics[width=0.46\linewidth]{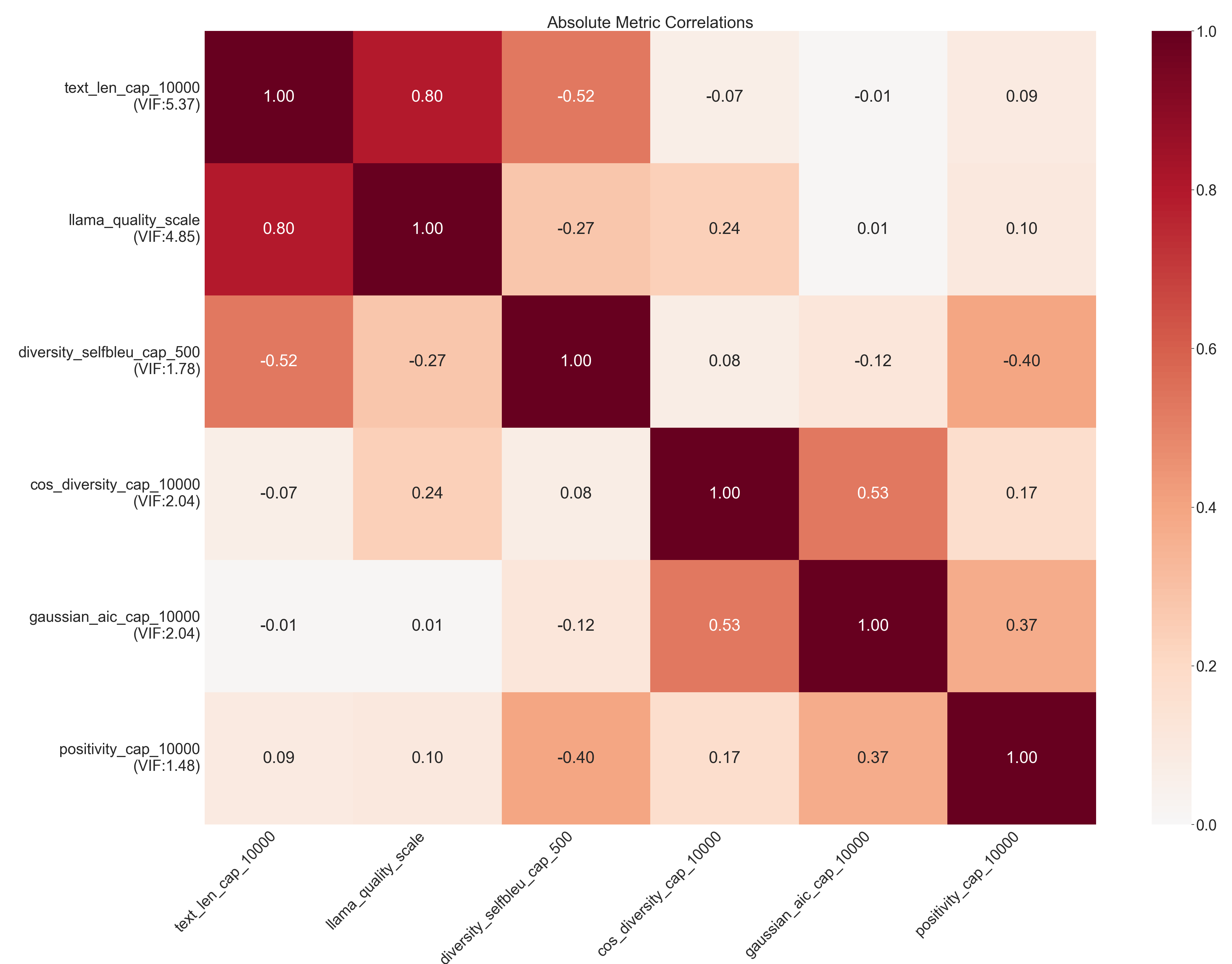}
\caption{\textbf{Variable Inflation Factor (VIF) scores before (left) and after (right) predictor variable selection}}
\label{fig:vif}
\end{figure*}

Figure \ref{fig:vif} shows Variance Inflation Factor (VIF) scores before and after predictor variable selection.
In the first step, we eliminated all predictors except \texttt{lexical\_entropy} (\texttt{word\_entropy}), \texttt{semantic\_entropy} (\texttt{kl\_entropy}), \texttt{quality}, \texttt{text\_length} (primarily to serve as a control for quality), \texttt{lexical\_diversity} (\texttt{diversity\_selfbleu}), \texttt{semantic diversity} (\texttt{cos\_diversity}), \texttt{Gaussianity}, and \texttt{Positivity}. 
In the first step, we kept the entropy metrics despite \texttt{kl\_entropy} having a relatively hight VIF score.
With those predictors, we conduct a pilot regression analyses (alike those in section \ref{sec:exp_reg}) experiment on four datasets from two domains: \webis, \redditsubmissions, \senatortweets, \twhm.
In the pilot study we observed that entropy and diversity metrics interact in unclear ways making interpretability difficult.
Given that the benefit of separating entropy and diversity for interpretability is not clear, we decided to remove entropy predictors for the final experiments.
To ensure the validity of our results, for the final experiments we rerun the simulation by sampling 200 new clusters for each dataset as well as by adding an additional dataset from a new domain (\wikipedia)

\subsection{Clustering}
\label{app:clustering}

To obtain data to be used for regression analysis, we create a number of subsets from each of the datasets.
This is done by the following procedure.
A 2D UMAP is fit on the 90k text embedded with \texttt{stella\_en\_1.5B\_v5} model \citep{stella}.
A series of clustering methods (dbscan, hdbscan, gmm, k-means) with different hyperparameters are done separately, each annotating the 90k samples.
Then the rest of the dataset is annotated by k-nn, with k=1.
This is done in two ways, with and without excluding the \textit{noise cluster} for the k-nn classifier.
This gives us a total of 120 different clusterings of the dataset.
Then for each clustering 10 clusters are taken, if there are not enough clusters over the size of 60k.
The remaining clusters are constructed by merging smaller clusters, either by uniformly sampling which cluster to merge or by iteratively merging the cluster that is the furthest away from the currently merged cluster. 
As this results in a large number of clusters (e.g. 1088 for \webis), we then obtain the final cluster set by subsampling 200 clusters.

To clarify, from the 200 clusters created for each dataset, some are created in a straight forward way (e.g. with k-nn) which will create the nice narrow clusters as the reviewer describes.
However, as discussed in this section, some clusters are created in a slightly more complex way, precisely to address this issue.
Some of the used clustering methods create very small clusters (miniclusters) which are then merged to create bigger final clusters to be used in the experiments.
This is sometimes done by randomly selecting the miniclusters to merge, and sometimes by iteratively adding the farthest away minicluster from the already merged miniclusters. Therefore, the final set of 200 clusters will be quite diverse.

\subsection{Fine-tuning procedure}
\label{app:ft}
We use the Unsloth library \citep{unsloth}(Apache license) to train model suing LoRA \citep{lora}.
The hyperparameters used are the default ones given by Unsloth, i.e. rank = 16, alpha = 16 batch size = 16, leaning rate = 2e-4, we use a linear schedules with 5 steps of warm-up.
For generation with use a temperature of 1.5 with min\_p \citep{minp} of 0.2.

\subsection{Iterative chain pseudocode}
Figure \ref{fig:iterative_chain_pseudocode} shows the pseudocode describing the iterative chain experiment used. Each iteration a fresh base model is selected and finetuned.

\begin{figure*}[t]
  \centering
  \begin{minipage}{0.95\textwidth}
\begin{lstlisting}[
  basicstyle=\ttfamily\scriptsize,
  numbers=left,
  numberstyle=\tiny,
  frame=single,
  framerule=0.6pt,
  framesep=6pt,
  columns=fullflexible,
  keepspaces=true
]
function iterative_chain(human_data, ai_ratio):
  base_models = [llama-3.2-1B, Qwen2.5-1.5B, SmolLM-1.7B, Falcon3-1B-Base]
  accumulated_data_pool = []
  for i in 20:
    iteration_base_model = sample(pretrained_models)

    if i == 0:
      training_set = sample(human_data, 8000)
    else:
      training_set = sample(accumulated_data_pool, 4000)

    iteration_ft_model = iteration_base_model.train_with_lora(training_set)

    iteration_new_data = iteration_ft_model.generate(4000*ai_ratio)
    iteration_human_data = human_dataset.sample(4000*(1-ai_ratio))

    accumulated_data_pool.add(iteration_new_data)
    accumulated_data_pool.add(iteration_human_data)
\end{lstlisting}
  \end{minipage}
  \caption{\textbf{Iterative chain pseudocode}}
  \label{fig:iterative_chain_pseudocode}
\end{figure*}

\subsection{Discussion on corrections for multiple comparisons}
We report uncorrected p-values in our regression analyses as our goal is not to test specific hypotheses about individual coefficients, but rather to identify general patterns of robust influence across datasets and conditions.
On a related note, variable selection was conducted prior to these analyses and applied to newly collected data (last sentence in Appendix \ref{app:feature_selection}).
We therefore believe that the risk of multiple testing bias is minimal.

In other words, we address the issue of Type I errors in the way we interpret our results and make conclusions (i.e. in a way that is robust to potential Type I errors): the fact that we consistently find the same predictors to be significant across different experiments gives a high confidence that those are not statistical artifacts.

\section{Additional results}

\subsection{Increasing the number of models per generation}
\begin{figure*}[ht!]
\includegraphics[width=\linewidth]{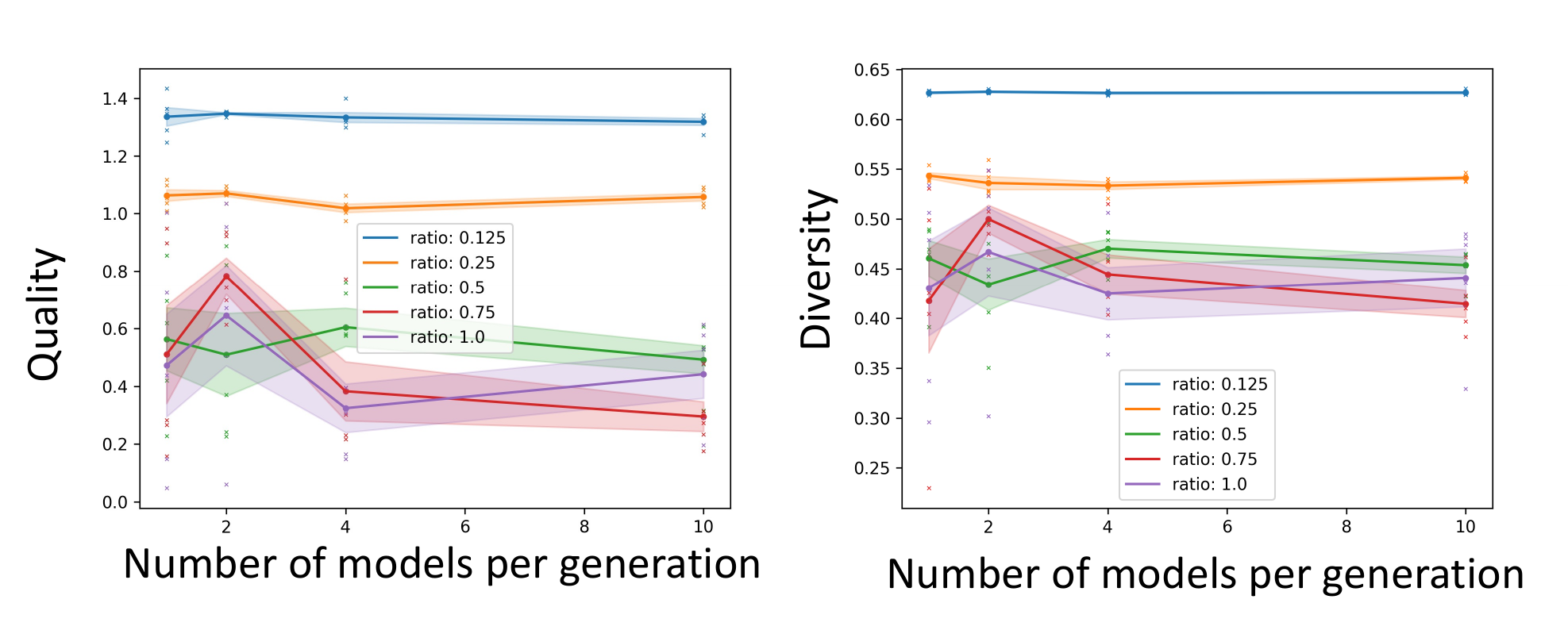}
\caption{\textbf{The effect of synthetic data ratio on shifts in diversity and quality holds when increasing the number of models per generation.} 
}
\label{fig:more_participants}
\end{figure*}

In the main text, we adopted the same experimental design as previous studies, where at each generation a single model is fine-tuned and used to generate new data. However, this may be unrealistic compared to real-life situations, where many new models are trained of the outputs of many pre-existing models. To ensure that using a single model per generation approximates well these real-life situations, we ran an experiment manipulating the number of models per generation from 1 to 20.
In Figure \ref{fig:more_participants}), we observe that increasing the number of models per generation does not qualitatively change the conclusion about the effect of synthetic data ratio on distribution shifts with respect to quality and diversity. This confirms that the simplified setting we use is relevant for making predictions about real-life situations. 

As a note, this pilot experiment was conducted with a slightly different quality metric, which ranks quality as either 0, 1 or 2. Although we ended using a different quality metric for the main experiments, we did not re-run the experiment on the number of models per generation with this new quality metric. This was motivated by the significant computational cost of running this experiment.  

\subsection{The effect of manipulating dataset quality on the distribution shift dynamics}

In this experiment, we explore the hypothesis that data quality is one of the potential factors influencing distribution shifts.
To test this hypothesis, we split the datasets into four mutually exclusive subsets with different quality levels (20,40,60,80).
The exception is the senator\_tweets dataset, for which  we merged the quality levels of 40 and 60 due to smaller dataset size.
Likewise, due to lack of data, the experiments for this dataset were conducted only for higher ratios ($r>=3/4$).
Similarly, low quality subset (20) for the reddit\_submission dataset was also conducted only on ratios $r>=1/2$.
We conduct this experiment of four datasets: \webis, \twhm, \senatortweets, and \redditsubmissions. 

\subsubsection{Effect on quality}
\label{sec:exp_quality}

\begin{figure*}[t!]
    \centering 
    \includegraphics[width=0.99\linewidth]{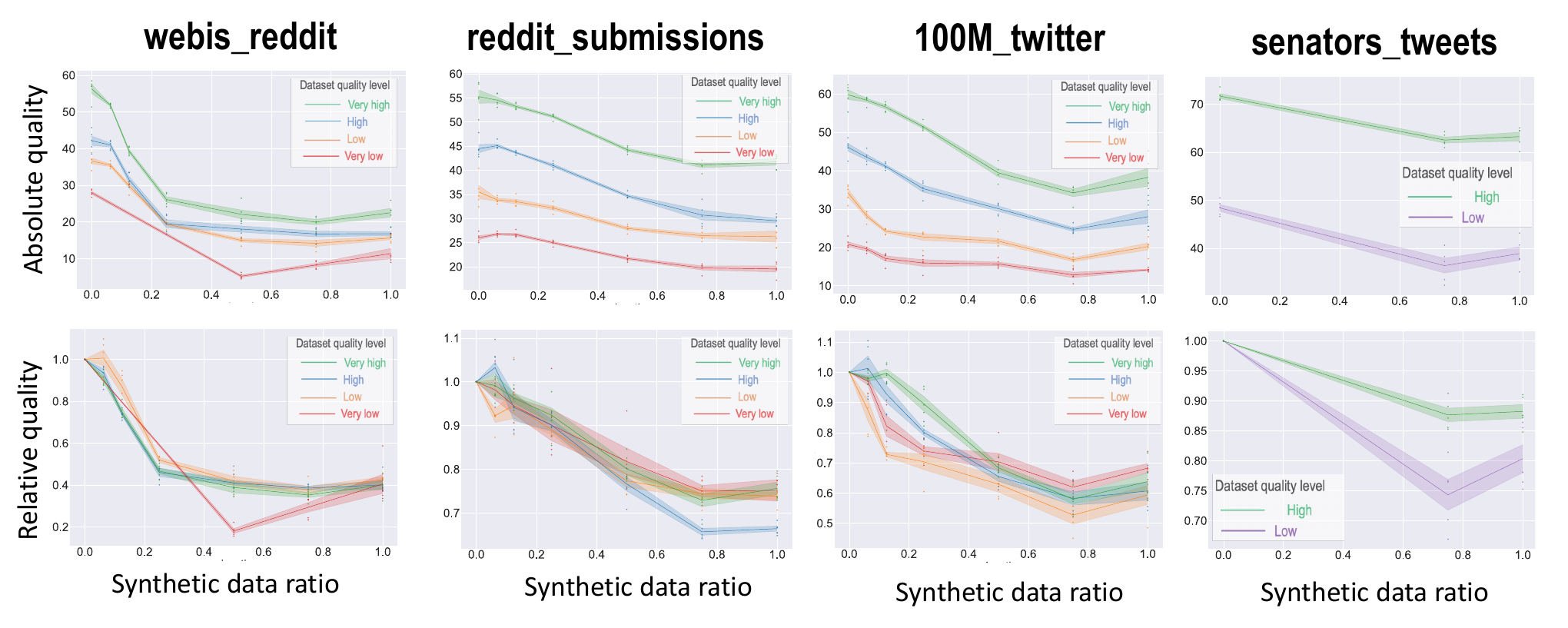}
    \caption{
    \textbf{Effect of human data quality on the \textit{rate} of degradation and sensitivity to synthetic data} Absolute measures (top row) correspond to the value of the corresponding metric at generation 19.
    Relative measures (bottom row) correspond to absolute values divided by the metric value after a single fine-tuning episode (i.e. generation 0).
    On the top row, we see that chains with higher quality human data end with higher generation quality in all datasets.
    On the bottom row, for the two Reddit datasets (third and fourth columns), we see that high quality chains also exhibit lower \textit{rates} of quality degradation and lower sensitivity to synthetic data (drops occur at higher synthetic data ratios).
    }
\label{fig:q_quality}
\end{figure*}

Figure \ref{fig:q_quality} shows the quality values of iterative chains for different quality levels.
As in section \ref{sec:exp_datasets}, we show the final absolute and relative quality levels as a function of synthetic data ratio.
Looking at \textit{absolute quality levels} (top row), we observe that, as expected, higher quality datasets also end with higher quality in the final generations.
More interestingly, looking at the \textit{relative quality levels}
of twitter datasets (bottom row), we observe that higher quality datasets lead to lower \textit{rate} of quality loss (distribution shift).
That is, not only does higher input data quality increase the quality of the final generated dataset, it also decreases the percentage of original quality lost due to recursive training.
Furthermore, focusing on lower ratios ($<1/4$) of the 100M\_tweets dataset, we observe that the higher quality dataset are more \textit{robust} to increasing synthetic data ratio.
For the quality of $80$, major quality losses are observed only at $r=1/4$, while for lower qualities it is observed already at $r=1/8$.
Finally, it should also be noted that the 100M\_tweets dataset with quality $20$ does not appear to lead to significant shifts with higher synthetic data ratios.
Given that the very low starting quality of this dataset, we believe that this is likely due to a \textit{floor} effect (i.e. there are no losses in quality because the initial dataset was already close to the lower bound.
Curiously, on both Reddit datasets, we do not observe strong differences when manipulating quality, implying that there are other factors than quality influencing collapse.
A similar study focusing on losses of diversity is presented in Appendix \ref{app:quality_div}, where an effect is observed only on the \senatortweets dataset.
Overall, this experiment shows that, in some conditions, high-quality datasets lead to increased robustness to distribution shifts.

\subsubsection{Effect on diversity}
\label{app:quality_div}

\begin{figure*}[ht!]
    \includegraphics[width=\linewidth]{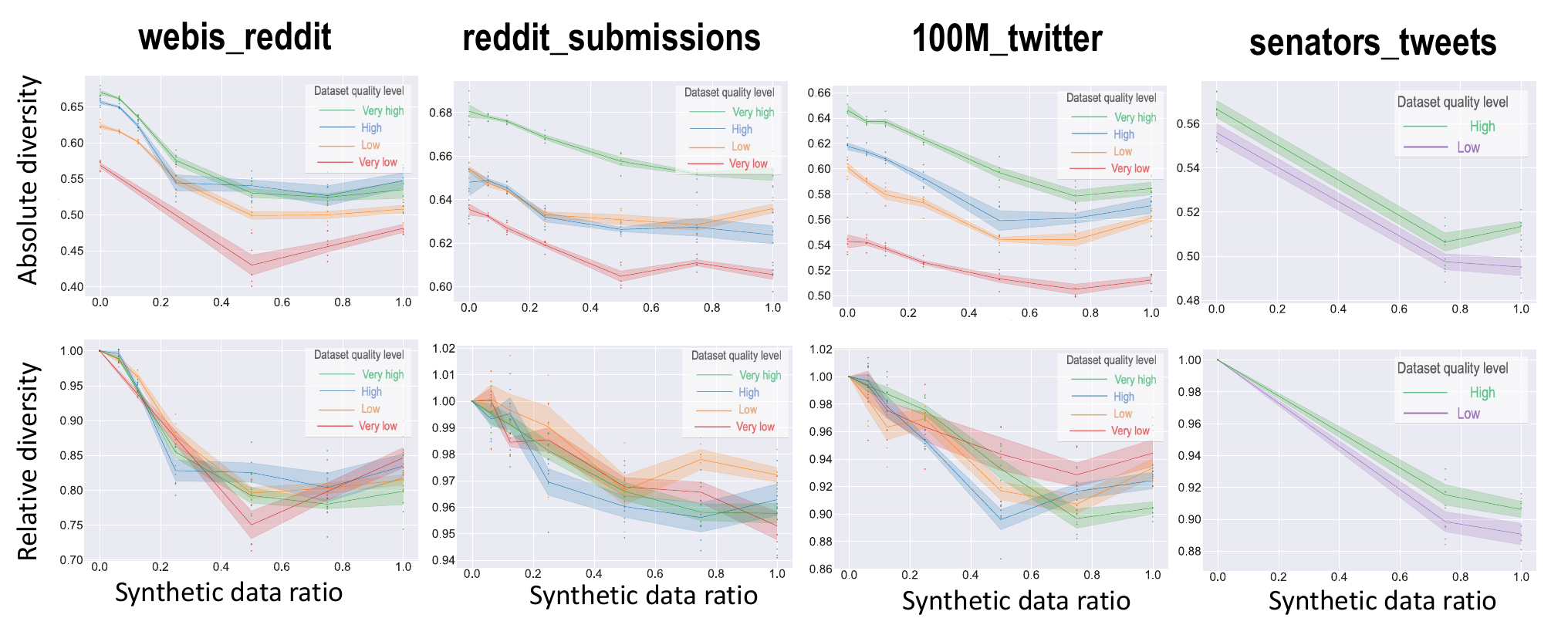}
    \label{fig:sub_d_pc}
\caption{\textbf{Effect of manipulating dataset quality on sensitivity to synthetic data ratio, for four different datasets.} Absolute measures correspond to the value of the corresponding metric at generation 19. Relative measure correspond to absolute values divided by the metric value after a single fine-tuning episode (i.e. generation 0).No clear effect is observed, except potentially on the \senatortweets dataset.
}
\label{fig:manip_quality_diversity}
\end{figure*}

Figure \ref{fig:manip_quality_diversity} shows the semantic diversity values of iterative chains for different quality levels.
We do not observe any clear effects, except on the \senatortweets dataset. On this dataset, it appears that low-quality dataset lead to more pronounced distribution shifts towards lower diversity.

\subsection{Toy model exploring the causes of the non-linear relationship between diversity loss and synthetic-data ratio}
\label{app:u_shape}

In section \ref{sec:exp_datasets}, we observed that for the Wikipedia dataset, the relationship between diversity loss and synthetic-data ratio was non-linear. Indeed, the greatest drops in diversity are observed for intermediate synthetic-data ratios, rather than for high values as in other datasets and previous works \cite{Bertrand2023, Bohacek2023, Kazdan2024}.
While at first surprising, we believe this pattern can be explained if we assume 
that synthetic data aligns more with the models' priors. This is not a strong assumption given that that data was generated by other fine-tuned versions of the same base models.
The intuition, which we experimentally confirm below, is that those datapoints aligned with the model's priors, have a stronger effect on the training process. And this then leads to the intermediate synthetic data ratios to essentially learn from less data.
Let us consider the three different synthetic-data ratios:
\begin{itemize}
    \item When synthetic-data ratio is low: While the model preferentially learns from synthetic data, most of its training data is human generated. Therefore, the model learns a distribution that resembles the human data despite this bias.
    \item When synthetic-data ratio is high: the model preferentially learns from synthetic data, but anyway most of the training data is synthetic. This bias thus does not have a large effect.
    \item When synthetic-data ratio is intermediate: the model preferentially learns from synthetic data, and receives human data and synthetic in comparable proportion. However, the bias will lead the model to essentially discard human data, and to learn only from the synthetic data, just like for high synthetic-data ratios. The difference is that the pool of synthetic data to learn from is here lower than in the high synthetic-data ratio. 
    
\end{itemize}

To test this hypothesis, we develop a toy example where we could manipulate whether learning from synthetic data aligned with models' priors is favored.
In this model, the true (\textit{human}) distribution is a uniform distribution over integers in $[0,N]$.
The \textit{model} is implemented as a normalized histogram over training datapoints, which are a combination of true and synthetic datapoints from the previous generations.
We sample N points from the true distribution, and normalize the resulting histogram to get the first model.
Then, we sample $r * N$ points from this model, and $(1-r) * N$ points from the true distribution.
We again derive a probability distribution from this sample to get the new model.
We repeat this process for 20 time steps. 

To introduce the bias mentioned in our hypothesis, we assume that the models have a prior to sample multiples of 2.
We thus multiply the histogram by a corresponding bias vector before normalizing.
Additionally, we can manipulate whether the human data overlaps with this bias: we can modify the true distribution so that it does not contain multiples of 2. 

\begin{figure*}[ht!]
    \includegraphics[width=\linewidth]{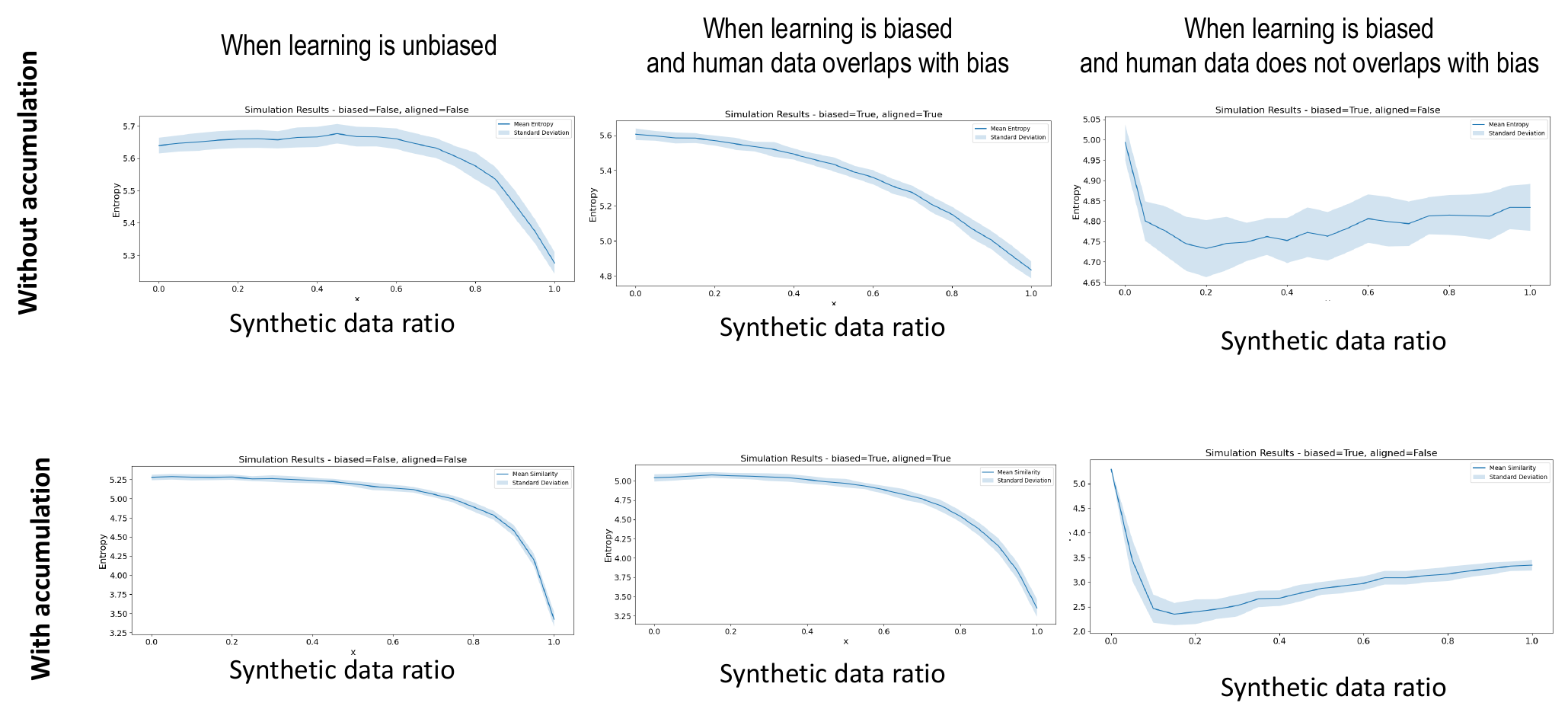}
    \caption{\textbf{Simulation results of the toy model.} Under specific conditions, a non-linear u-shaped relationship between diversity loss and synthetic-data ratio emerges.}
\label{fig:toy_model}

\end{figure*}

We then ran 50 simulations for different values of synthetic-data ratio, manipulating whether learning is biased and whether the true distribution overlaps with this bias. As show in Figure \ref{fig:toy_model}, we observe that when learning is biased and the true distribution does not overlap (right column), we can reproduce the U-shape found for the Wikipedia dataset. This non-linear relationship disappears when we remove this bias (left column) or when the true distribution overlaps with this bias. We were able to observe this pattern both in the Accumulation (top-row) and no-Accumulation (bottom row) settings.
This is therefore consistent with our hypothesis for explaining the observed U-shape relationship. 

\subsection{Regression models' explained variances}
Here we provides $r^2$ measures for the linear regression models used in experiments in sections \ref{sec:exp_reg} and \ref{sec:exp_reg_merged}.
Table \ref{tab:r2_reg} corresponds to table \ref{tab:reg} from section \ref{sec:exp_reg}.
Table \ref{tab:r2_merged} corresponds to Figure \ref{fig:merged} from section \ref{sec:exp_reg_merged}.
We can see that the considered properties can partially account for observed effect, many more effects remain to be uncovered (in particular in the 100M tweets and wikipedia datasets).

\begin{table*}[t]
\centering
\small
\begin{tabular}{lccccccccc}
\hline
$\mathbf{r^2}$ & All & \multicolumn{2}{c}{Webis} & \multicolumn{2}{c}{100M Tweets} & \multicolumn{2}{c}{Reddit Submissions} & \multicolumn{2}{c}{Wikipedia} \\
\hline
\textbf{Synthetic data ratio} &  & 1/8 & 1/4 & 1/8 & 1/4 & 1/8 & 1/4 & 1/8 & 1/4 \\
\hline
\textbf{Semantic Diversity} & 0.293 & 0.244 & 0.541 & 0.043 & 0.188 & 0.109 & 0.502 & 0.147 & 0.105 \\
\textbf{Quality}            & 0.737 & 0.482 & 0.353 & 0.193 & 0.324 & 0.143 & 0.564 & 0.249 & 0.276 \\
\hline
\end{tabular}
\caption{$R^2$ results corresponding to Table \ref{tab:reg}.}
\label{tab:r2_reg}
\end{table*}

\begin{table}[t]
\centering
\small
\begin{tabular}{lcc}
\hline
\textbf{Domain} & $R^2$ (Diversity) & $R^2$ (Quality) \\
\hline
\textbf{Webis}       & 0.340 & 0.591 \\
\textbf{100M Tweets} & 0.532 & 0.611 \\
\textbf{Wikipedia}   & 0.241 & 0.363 \\
\hline
\end{tabular}
\caption{$R^2$ results corresponding to Figure \ref{fig:merged}.}
\label{tab:r2_merged}
\end{table}

\subsection{Additional experiments on the distribution shift of political lean}
\label{app:pol}

\begin{figure}
\begin{subfigure}{0.44\linewidth}
    \includegraphics[width=\linewidth]{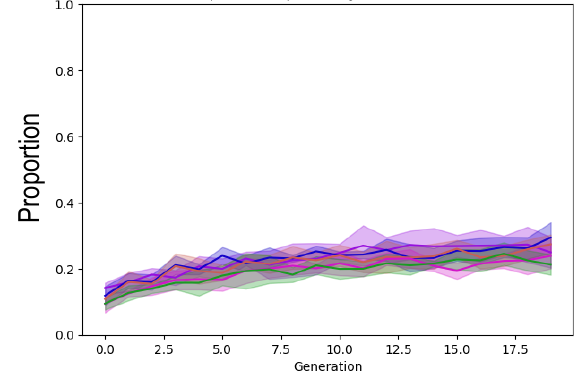}
    \caption{Proportion of politically neutral tweets}
    \label{fig:neutral_evolution}
\end{subfigure}
\hfill
\begin{subfigure}{0.44\linewidth}
    \includegraphics[width=\linewidth]{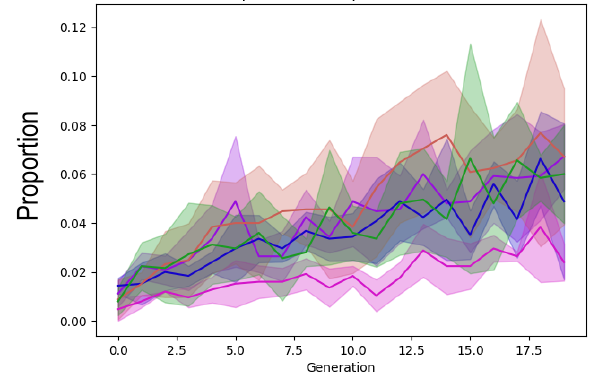}
    \caption{Proportion of non-political tweets}
\label{fig:non_pol_evolution}
\end{subfigure}
\caption{
(a) Proportion of politically neutral tweets increases, implying that models tend to avoid strong political statements.
(b) Proportion of non-political tweets marginally increases, implying that the models stay on the topic of politics.
}
\end{figure}

We provide here the additional figures that are discussed in section \ref{sec:pol}.
On Figure \ref{fig:neutral_evolution} we observe a steady increase in the proportion of  ``perfectly neutral'' tweets (with assigned a score of exactly 50).
On Figure \ref{fig:non_pol_evolution} while we observe a slight increase in the number of non-political tweets, those remain marginal, indicating that the models are able to maintain the focus on political topics (Figure \ref{fig:neutral_evolution}). 
This suggests that generated tweets remain in the topic of politics, but drift from strong partisanship.

\begin{figure}
    \centering
    \includegraphics[width=0.8\linewidth]{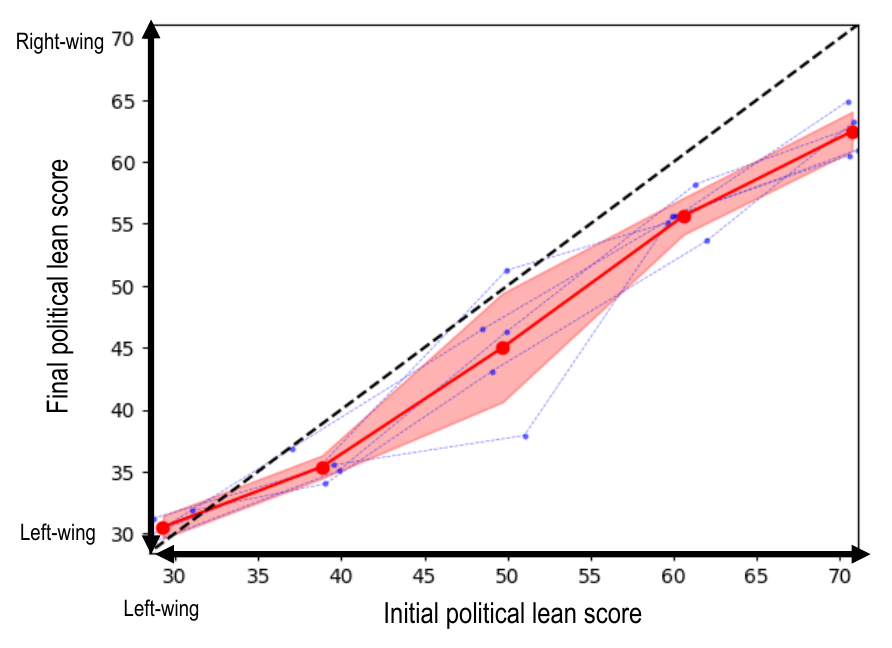}
\caption{Average political lean at the last generation as a function of political lean in the true distribution after excluding political tweets}
\label{fig:lean_in_out_exclude}
\end{figure}

Figure \ref{fig:neutral_evolution} revealed that the shift is partly driven by an increase in politically neutral content. To isolate the different mechanisms at play, we performed the same analysis, but without taking politically neutral tweets into account (Figure \ref{fig:lean_in_out_exclude}).
The consequence of this manipulation was to accentuate the observed asymmetry, as the political lean that minimizes shift magnitude moves even more toward the left.
This suggests that there might two interacting mechanisms influencing political lean evolution: first, a tendency to generate politically-neutral content; and a tendency to shift the distribution toward left-wing content.

\begin{figure}
    \centering
    \includegraphics[width=0.8\linewidth]{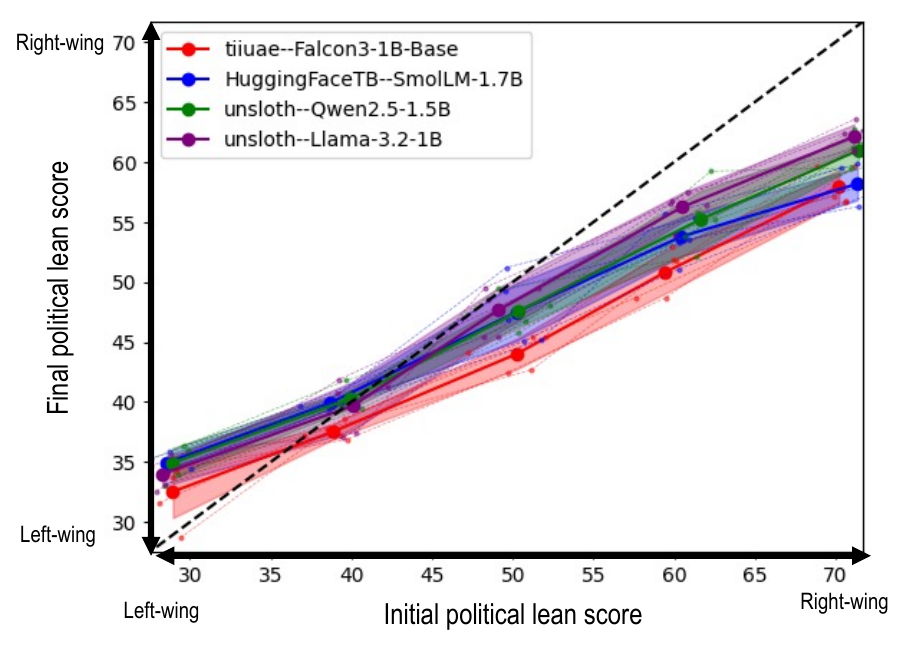}
\caption{Average political lean at the last generation as a function of political lean in the human distribution with homogeneous transmission chains. We observe slight differences between chain: for instance, Falcon3-1B chain appears to have a stronger left-wing bias than others, while it is weaker for Llama-3.2-1B}
    \label{fig:lean_homogeneous}
\end{figure}

Experiments in the paper were conducted with mixed-model iterative chains - each generation a fresh base model is sampled out of four possible models options (\texttt{LLama-3.2-1B}, \texttt{Qwen2.5-1.5B}, \texttt{SmolLM-1.7B}, \texttt{Falcon3-1B-Base}).
As different models may display different biases with respect to political lean, we also ran an experiment with homogeneous chains, where the same base model is used in each generation.
To clarify, in each generation the training still starts from a new instance of a pretrained model, e.g. over 20 generations we will initialize 20 separate LLama-3.2-1B instances.
Figure \ref{fig:lean_homogeneous} compares homogenous chains corresponding to the four considered models.
We observe little variation between the chains, although Falcon3-1B seem to display slightly more pronounced left-wing bias, while this bias is weaker for Llama-3.2-1B.

\end{document}